\documentclass[a4paper,11pt]{article}

\usepackage[utf8]{inputenc}

\usepackage{quoting}

\usepackage{amsmath, amsfonts, amssymb}
\usepackage[amsmath,thmmarks]{ntheorem}
\usepackage[mathcal]{euscript}
\usepackage{bm,bbm}
\usepackage{multirow}

\usepackage[ruled,vlined,linesnumbered]{algorithm2e}
\usepackage{algpseudocode}

\usepackage{caption}
\usepackage[table]{xcolor}
\usepackage{array}

\usepackage{graphicx}
\usepackage{subfig}
\usepackage{enumitem}
\usepackage{url}

\usepackage{acronym}

\newcommand{\etal}{\emph{et al.} }

\newcommand{\fref}[1]{Figure~\ref{#1}}

\newcommand{\tref}[1]{Table~\ref{#1}}
\newcommand{\sref}[1]{Section~\ref{#1}}
\newcommand{\algoref}[1]{Algorithm~\ref{#1}}

\newcommand{\db}[1]{\texttt{#1}}

\newcommand{\p}{\hphantom{1}}

\theoremstyle{plain}
\theoremheaderfont{\normalfont\bfseries}\theorembodyfont{\slshape}
\theoremseparator{:}

\usepackage[affil-it]{authblk}

\usepackage{hyperref}
\begin{document}

\acrodef{AE}{AutoEncoder}
\acrodef{ANN}{Artificial Neural Network}
\acrodef{CV}{Computer Vision}
\acrodef{CNN}{Convolutional Neural Network}
\acrodefplural{CNN}[CNNs]{Convolutional Neural Networks}
\acrodef{CSU}{Cellular Structures and Ultrastructure}
\acrodef{DoG}{Difference of Gaussian}
\acrodef{DAE}{Denoising AutoEncoder}
\acrodefplural{DAE}[DAEs]{Denoising AutoEncoders}
\acrodef{DL}{Deep Learning}
\acrodef{DNN}{Deep Neural Network}
\acrodefplural{DNN}[DNNs]{Deep Neural Networks}
\acrodef{EM}{Electron Microscopy}
\acrodef{LoG}{Laplacian of Gaussian}
\acrodef{NN}{Neural Network}
\acrodefplural{NN}[NNs]{Neural Networks}
\acrodef{ML}{Machine Learning}
\acrodef{MLP}{Multi-Layer Perceptron}
\acrodef{ROC}{Receiver Operating Characteristic}
\acrodef{SAE}{Stacked Autoencoder}
\acrodefplural{SAE}[SAEs]{Stacked Autoencoders}
\acrodef{SDA}{Stacked Denoising Autoencoder}
\acrodefplural{SDA}[SDAs]{Stacked Denoising Autoencoders}
\acrodef{SD}{Spot Detector}
\acrodef{SVM}{Support Vector Machine}
\acrodef{TEM}{Transmission Electron Microscopy}
\acrodef{TEMicroscope}{Transmission Electron Microscope}
\acrodef{TL}{Transfer Learning}

\title{Stacked Denoising Autoencoders and Transfer Learning for Immunogold Particles Detection and Recognition}

\author{Ricardo~Gamelas~Sousa}
\affil{Farfetch, Porto, Portugal
}

\author{Jorge~M.~Santos}
\affil{Instituto Superior de Engenharia, Polit\'{e}cnico do Porto, Portugal
}

\author{Lu\'{i}s~M.~Silva}
\affil{Dep. de Matem\'{a}tica at Universidade de Aveiro, Portugal 
}

\author{Lu\'{i}s~A.~Alexandre}
\affil{Instituto de Telecomunica\c{c}\~{o}es, Universidade da Beira Interior, Rua Marqu\^es d'\'Avila e Bolama, 6201-001, Covilh\~a, Portugal
}

\author{Tiago Esteves} 
\affil{Instituto de Investiga\c{c}\'{a}o e Inova\c{c}\'{a}o em Sa\'{u}de (i3S) Porto, Portugal}

\author{Sara Rocha}
\affil{Centro de Biotecnologia dos A\c{c}ores (CBA), Universidade dos A\c{c}ores, Portugal}

\author{Paulo Monjardino}
\affil{Centro de Biotecnologia dos A\c{c}ores (CBA), Universidade dos A\c{c}ores, Portugal}

\author{Joaquim~Marques~de~S\'{a}}
\affil{Instituto de Engenharia Biom\'{e}dica (INEB), Porto, Portugal}

\author{Francisco Figueiredo}
\affil{Instituto de Investiga\c{c}\'{a}o e Inova\c{c}\'{a}o em Sa\'{u}de (i3S) Porto, Portugal}

\author{Pedro Quelhas}
\affil{Instituto de Investiga\c{c}\'{a}o e Inova\c{c}\'{a}o em Sa\'{u}de (i3S) Porto, Portugal}

\maketitle
\begin{abstract}
In this paper we present a system for the detection of immunogold particles and a \ac{TL} framework for the recognition of these immunogold particles. Immunogold particles are part of a high-magnification method for the selective localization of biological molecules at the subcellular level only visible through \ac{EM}. The number of immunogold particles in the cell walls allows the assessment of the differences in their compositions providing a tool to analise the quality of different plants. For its quantization one requires a laborious manual labeling (or annotation) of images containing hundreds of particles. The system that is proposed in this paper can leverage significantly the burden of this manual task.

For particle detection we use a \ac{LoG} filter coupled with a \ac{SDA}. In order to improve the recognition, we also study the applicability of \ac{TL} settings for immunogold recognition. \ac{TL} reuses the learning model of a source problem on other datasets (target problems) containing particles of different sizes.
The proposed system was developed to solve a particular problem on maize cells, namely to determine the composition of cell wall ingrowths in endosperm transfer cells. This novel dataset as well as the code for reproducing our experiments is made publicly available.

We determined that the \ac{LoG} detector alone attained more than 84\% of accuracy with the F-measure. Developing immunogold recognition with \ac{TL} also provided superior performance when compared with the baseline models augmenting the accuracy rates by 10\%.
\end{abstract}


\section{Introduction}\acresetall
Immunogold electron microscopy is a high-magnification method for the selective localization of biological molecules at the subcellular level. Antibodies coupled to particles of colloidal gold, which are visible in the transmission electron microscope, can reveal the localization and distribution of biological molecules of interest. In this particular work, this technique was used to determine the composition of cell wall ingrowths of maize (\emph{Zea mays} L.) endosperm transfer cells~\cite{monjardino2013}. Wall ingrowths are uneven thickenings of the cell wall that increase significantly the area of the plasmalemma, thus enabling to sustain a high number of membrane transporters and produce better plants. In addition, INCW2, a cell wall-bound invertase and an essential enzyme for passive sugar flow into the endosperm, is mostly located in the wall ingrowths~\cite{Kang2009}. Therefore, wall ingrowths contribute significantly to enhance transport capacity of assimilates into the endosperm, thus having a significant impact on kernel yield.

Most basal maize endosperm transfer cells differentiate reticulate ingrowths in the outer periclinal wall and flange ingrowths in the inner periclinal walls and anticlinal walls~\cite{monjardino2013}. This unique feature has not been reported in any other species and makes these cells a unique model to study the differences and similarities of both types of ingrowths and underlying primary cell wall.

\begin{figure*}[t]
  \begin{center}
    \includegraphics[width=.9\textwidth]{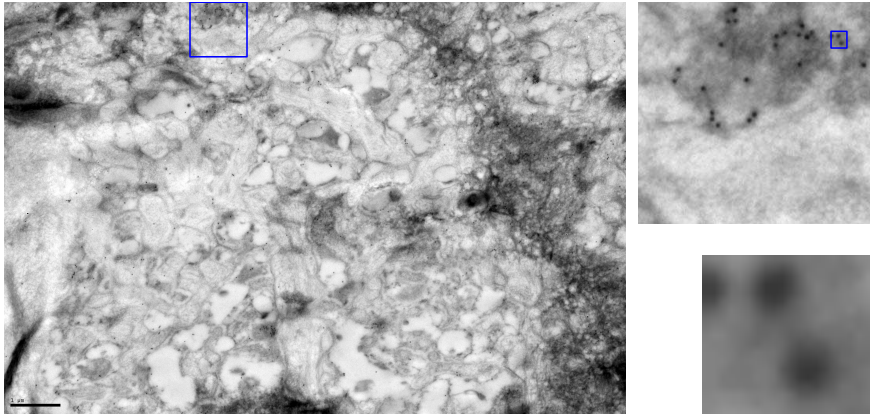}
    \caption{Representative images of our datasets illustrating different structures that can interfere in the recognition of the immunogold particles due to: cellular overlapping, tissues and background noise. Each image has $4008\times 2670$ pixels of dimension with particles' \emph{diameter} ranging from 8 to 20 pixels. In this particular image we have a sample with a magnification of \db{15000} (scale of $1\mu m$ to the side of the image) with particles with a diameter of 8 pixels. Right to the main figure, we have a magnification of a cluster of immunogold particles with noise and background; in the bottom a magnification of the cluster is also depicted in a $20 \times 20$ patch containing two particles.}
    \label{fig:dataset}
  \end{center}
\end{figure*}

In the process of analyzing the cell’s ultra-structure it is important to have different magnifications. A wider field of view is achieved with lower magnifications and a more detailed analysis of the cell’s ultra-structure is achieved with higher magnifications. In less magnified samples it is more difficult to detect the immunogold particles that are easily confused with cell’s ultra-structure, but enable the identification of the larger structures of the cells, namely the ingrowths and adjacent walls (see \fref{fig:dataset}). Higher magnification images provide a detailed analysis of the ultra-structure essential to this study (see also Figure 1 in~\cite{RGamelasSousaIWANN2015}).

The use of specific antibodies reveals the differences and similarities of components of the walls and ingrowths throughout the development of transfer cells. The automatic detection of immunogold particles is a valued tool to facilitate the calculation of the amount of immunogold particles in the walls and ingrowths studied and by that to determine the differences in their composition. One drawback is that the quantification of immunogold particles is a very time-consuming and prone to error task~\cite{ribeiro2006computer} which can benefit from an automatic detection and recognition tool.

Given an approximate size of the immunogold particles radius, we show that the \acf{LoG} filter is tolerant to shape variations and to noise that may occur during the image acquisition. The feasibility of this approach is however limited by the quality of the image --- see \fref{fig:dataset}. To overcome this constraint, we propose a framework that couples a \acf{SDA} to filter the detections of the \ac{LoG} filter and thus accelerate the whole process. This framework is tolerant to false detections by the \ac{LoG} filter and afterwards uses the capability of the \ac{SDA} to extract high representative features from our images towards the improvement of the detection rates~\cite{RGamelasSousaICIAR2015}.

Another proposal of this manuscript refers to how to handle datasets corresponding to immunogold particles captured with different magnifications: one simple way is to devise independent \ac{ML} techniques so we can analyze each one properly. \ac{TL} accelerates this process by \emph{reusing} a classifier designed for a given (\emph{source}) problem on another (\emph{target}) problem, with some similarities with the original one~\cite{RGamelasSousaIWANN2015}. The rationale to apply TL to immunogold particles detection is related to the difficulty to identify these structures, especially in lower magnifications, due to noise, complex structures and annotations with feeble quality. On the other hand, with higher magnifications the distinction between artifacts and structures can be conducted in a easier way due to clear boundaries.  For this reason, it is straightforward to explore the advantages of using TL by transferring a model that was obtained on a higher magnification to a dataset of a lower magnification.


The detection and recognition of \ac{CSU} in \ac{EM} is still in its infancy. We will still outline some of the few works that have been presented in the literature in \sref{sec:prevwork}. Our proposal for the detection of immunogold particles using our combined approach with the \ac{LoG} and \ac{SDA} will be described in \sref{sec:detection}. Since it is common to acquire samples in different magnifications we explore our hypothesis on the benefits of using \ac{TL} for immunogold recognition in \sref{sec:TL}. Experimental study and discussion of the results are presented in \sref{sec:results} and finally, conclusions are drawn in \sref{sec:conclusions}.

\section{Previous Works}
\label{sec:prevwork}
Automatic cellular structure and ultra-structure detection in microscopy imaging has been evolving fast in the literature, mostly due to requirements of processing massive data storages or the emergence of new microscopy technologies~\cite{knott2013}. For example, Fisker \etal in~\cite{fisker2000estimation} explore the possibility of automatically estimating particle sizes in immuno-microscopy imaging. Their approach is based on deformable models that can be fitted to the prior known shape of the particles. A different approach was presented by Mallic \etal in~\cite{mallick2004detecting} with the same goal as Fisker, where a cascade of classifiers was employed.

Beyond these works, Ribeiro \etal in~\cite{ribeiro2006computer} contributed with an insight review on the processing of images generated by electron microscopies and the advantages of methodologies. Obviously, experimental speed-ups and reproducibility are some of the most mentioned valuable aspects of these tools. In the limit, and since complex structures viewed in \ac{EM} acquisitions entail a manual analysis of such images by one (or several) experts (which  can take several hours or even years to conclude --- see page 7 in~\cite{knott2013}), this strengthens the motivation for (semi-) automated tools for \ac{EM} image analysis.

On cryo-electron microscopy images (a technology similar to \ac{EM}), Voss, Woolford and co-workers~\cite{voss2009dog,woolford2007laplacian} applied a \ac{DoG} and \ac{LoG} filters as a first step to detect biological structures. However, they were not tailored for immunogold particles. For the detection of biological structures, there is an algorithm entitled \ac{SD}~\cite{olivo2002extraction} that is included in the well-known Icy bioimaging software (Icy, in short)~\cite{de2012icy}. \ac{SD} is based on the non-decimated wavelet transform allowing the detection of spots that can be organelles or other biological structures~\cite{olivo2002extraction}. This approach aggregates a response for each magnification and scale of the image providing detailed information of the objects. A major drawback is concerned with the number of parameters required to function properly (e.g., the identification of a trade-off between particles and background; the definition of a scale and sensibility that controls both size of the particles to be detected and a threshold for noise removal), which may be infeasible for a non-expert user.
In a work more closely related to ours, Wang \etal~\cite{Wang2011} explored a \ac{DoG} approach, a computational cost optimization of \ac{LoG}, for the detection of immunogold particles on \ac{EM} images.

Note that automatic organelle detection and identification (segmentation and classification), especially in \ac{EM} imaging, is relatively recent (see for instance refs. Straehle et al., 2011; Kreshuk et al., 2011 in~\cite{knott2013}). For this reason, there are very few works that tackle explicitly the biomedical analysis on \ac{EM} images. For the segmentation of mitochondria an extension of the super-pixel approach was proposed in Fua \etal (see~\cite{fua2012} and ref. 17 in~\cite{fua2014}) or, as Ciresan \etal proposed using \acp{DNN}~\cite{ciresan2012neural,ciresan2012deep}. More recently, Huang and co-workers have tackled the detection of other cellular structures based on a symmetry transform~\cite{HuangMIUA2014}.

In this paper we will address the detection and recognition of immunogold particles. A major difference from the aforementioned proposals is that organelles are irregular (e.g., in shapes and intensities) and with distinct aspects between themselves. This work is focused on an intuitive tool for the detection of immunogold particles with regular spherical shape, thus avoiding the adoption of a highly parameterized formalism. Moreover, this proposal combines a conventional \ac{LoG} filter with a Machine Learning algorithm (Transfer Learning with SDAs) improving the results produced by the \ac{LoG} alone. With the nonexistent (to the best of our knowledge) of an equivalent approach, this work aims to provide a first approach to tackle complex biological structures existing in \ac{EM} imaging, thus paving the way to other robust machine learning methodologies.

\section{Immunogold Particles Detection}
\label{sec:detection}

We present a cooperative approach for particle detection and recognition. Our approach encompasses coupling the detection performed by a \ac{LoG} filter and a post-processing stage conducted by a deep learning approach (\ac{SDA}).

\subsection{Immunogold Particle Detection using a LoG Filter}

For the task of immunogold particle detection we used the LoG filter. This filter was first introduced by Lindeberg~\cite{Lindeberg} and is based on the image scale-space representation that allows for the detection of blob-like structures in images.

Given an input image $I(x,y)$, the Gaussian scale-space representation at a certain scale $t$ is represented as:
\begin{eqnarray}
  L(x,y,t) & = g(x,y,t) \ast I(x,y) \\
  g(x,y,t) & = \frac{1}{2\pi t} e^{-\frac{x^2 + y^2}{2t}}, \nonumber
\label{eq:scale_space}
\end{eqnarray}
where $\ast$ is the convolution operator. Given this, the scale normalized \ac{LoG} operator is defined as:
\begin{equation}
    \triangledown^2 L(x,y,t) = t^2 (L_{xx}(x,y,t) + L_{yy}(x,y,t)),
\label{eq:log}
\end{equation}
where $L_{xx}$ and $L_{yy}$ are the image second derivatives in $x$ and $y$ respectively, and $t$ is the scale parameter so that $t = r / 1.5 $ for a particle radius~$r$~\cite{TEstevesibpria2013,RGamelasSousaICIAR2015}.

\begin{figure*}[!ht]
\centering
\subfloat[]{\includegraphics[width=.25\textwidth]{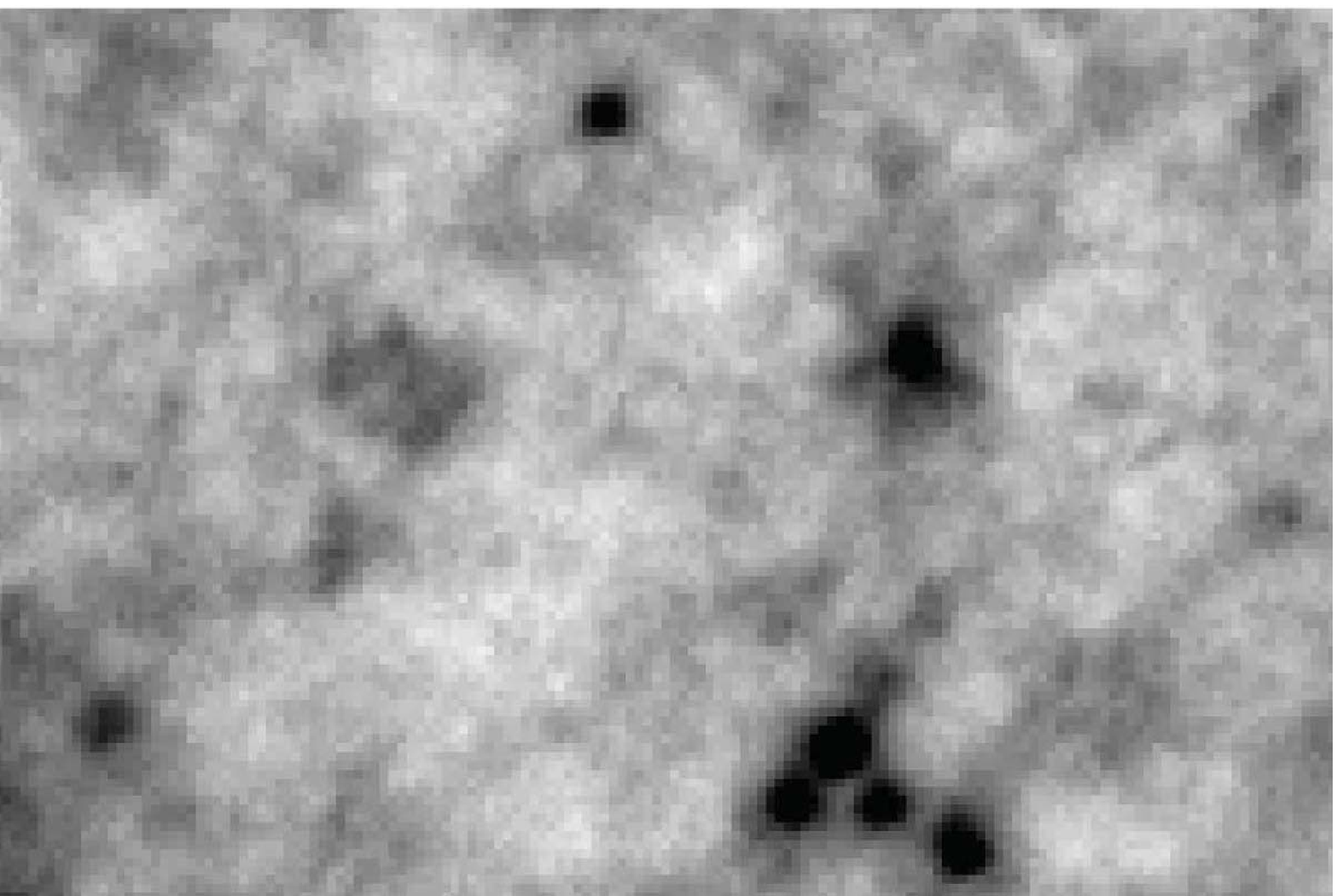}}
\subfloat[]{\includegraphics[width=.25\textwidth]{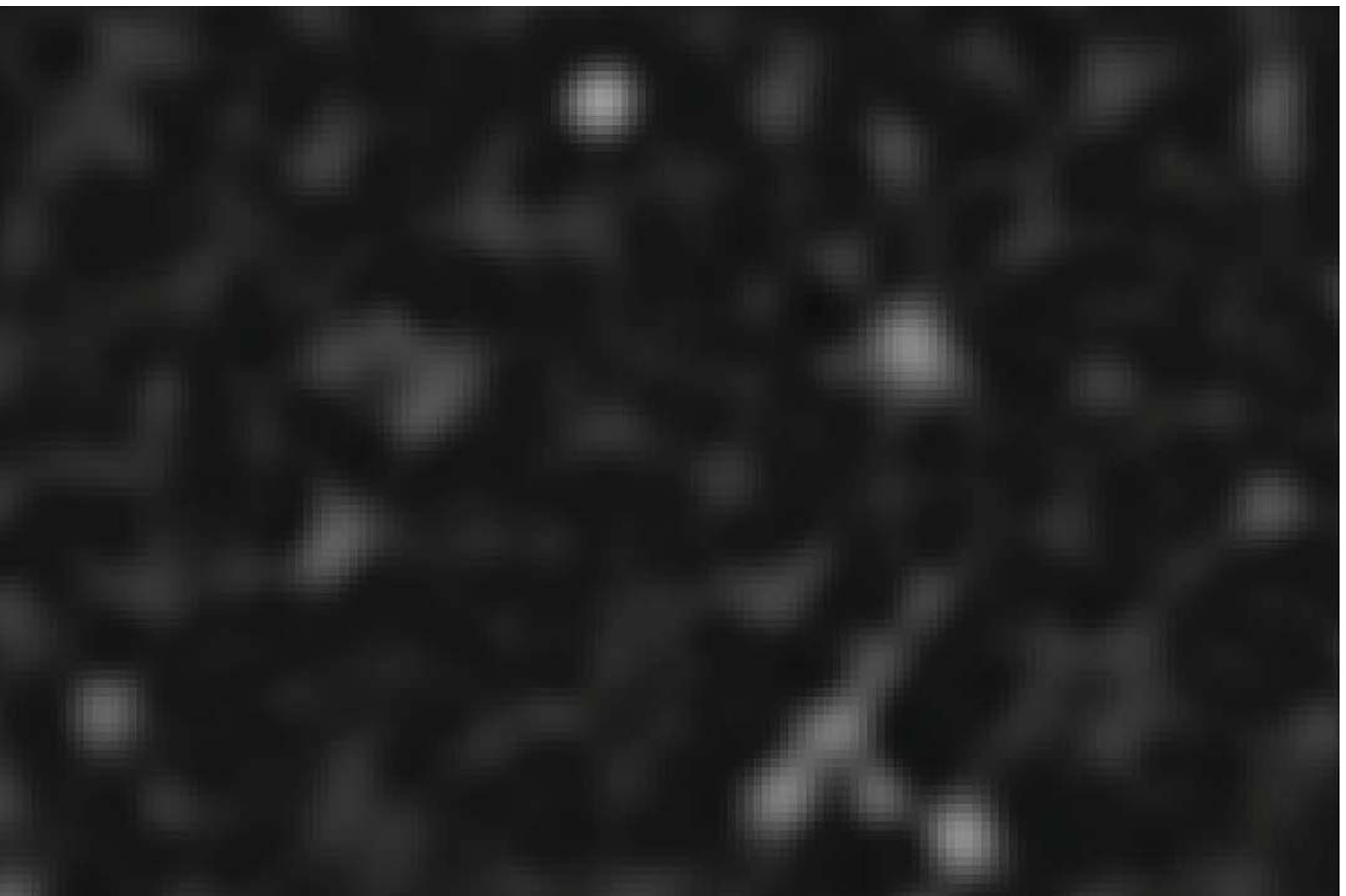}}
\subfloat[]{\includegraphics[width=.25\textwidth]{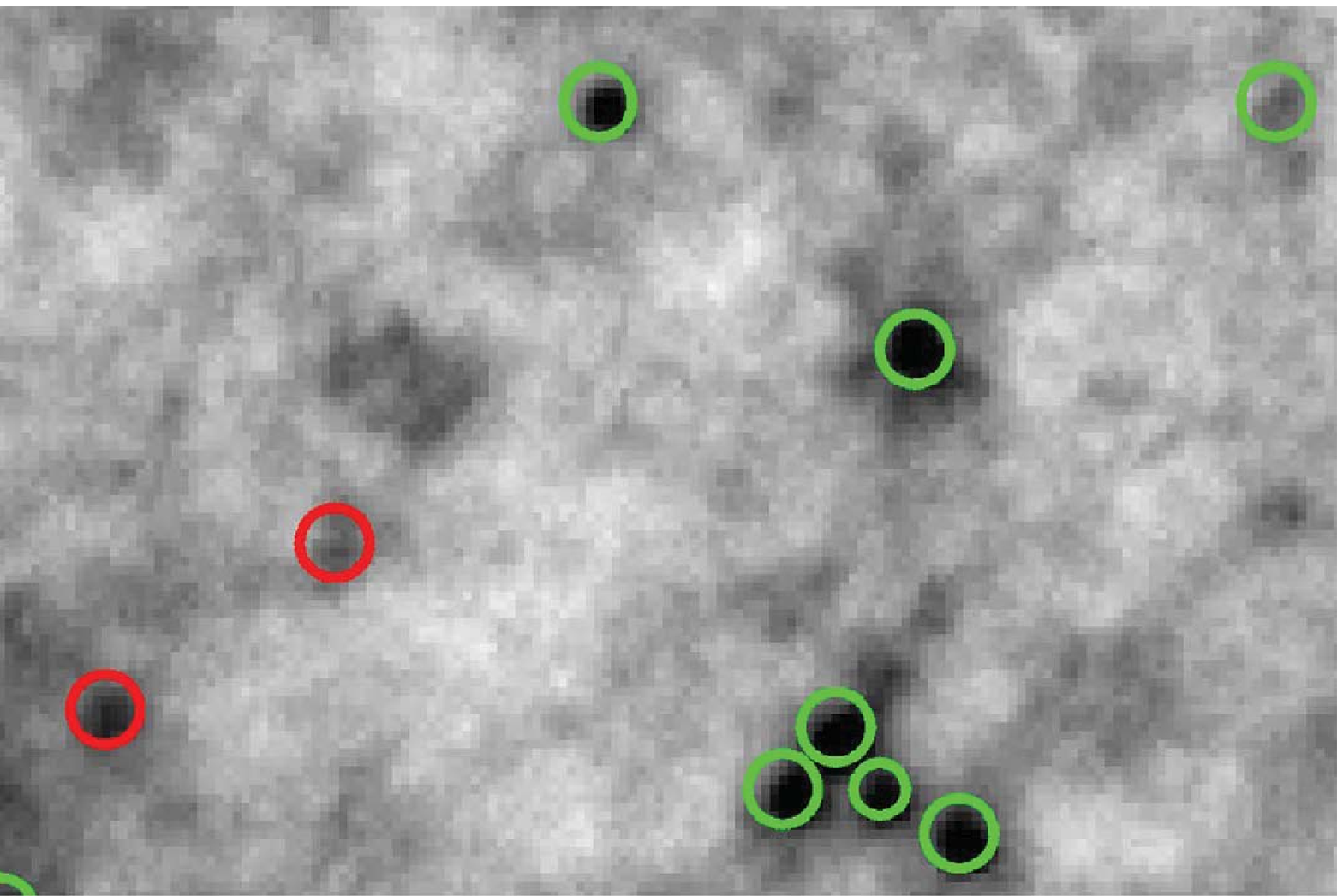}}

\subfloat[]{\includegraphics[width=.25\textwidth]{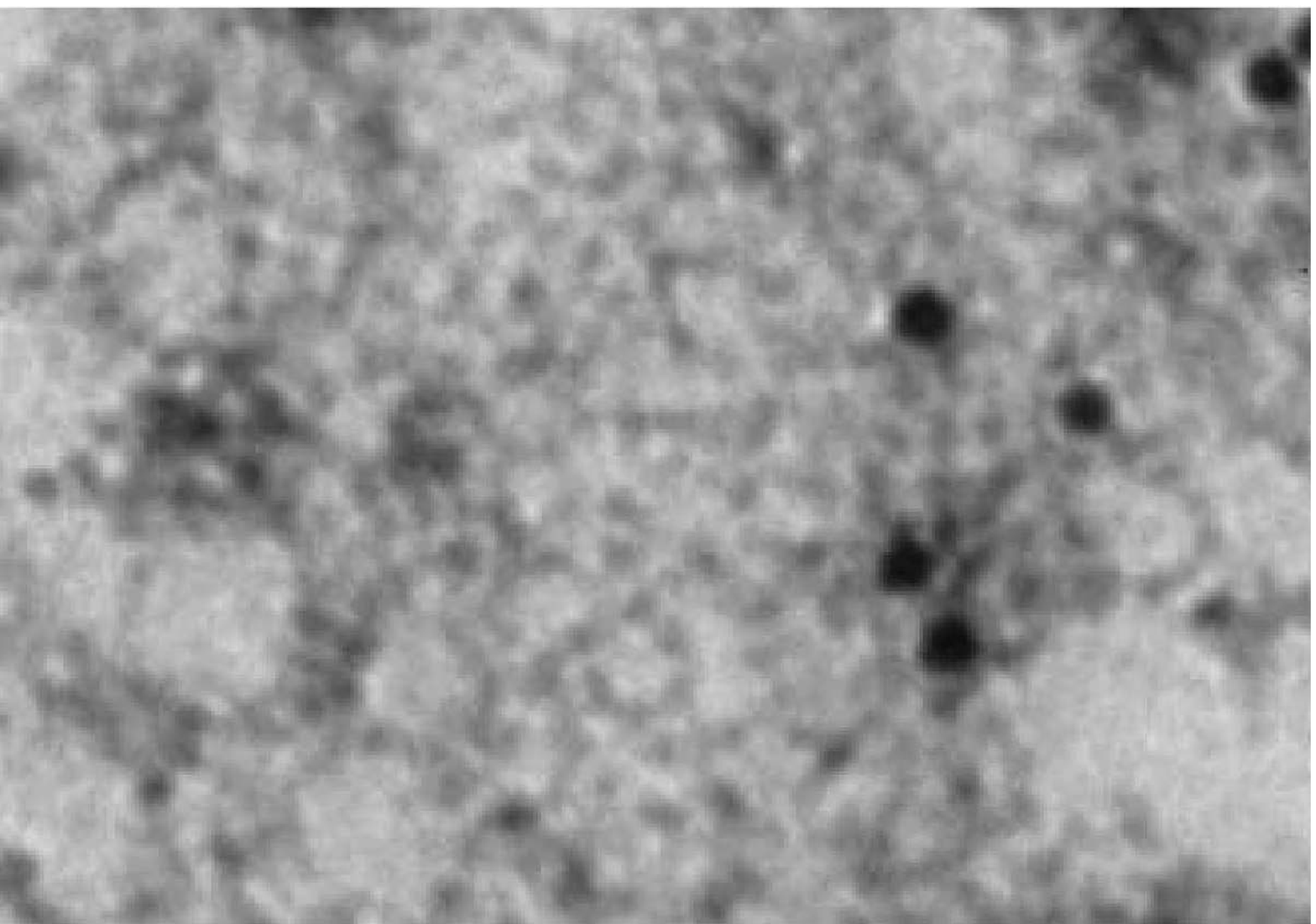}}
\subfloat[]{\includegraphics[width=.25\textwidth]{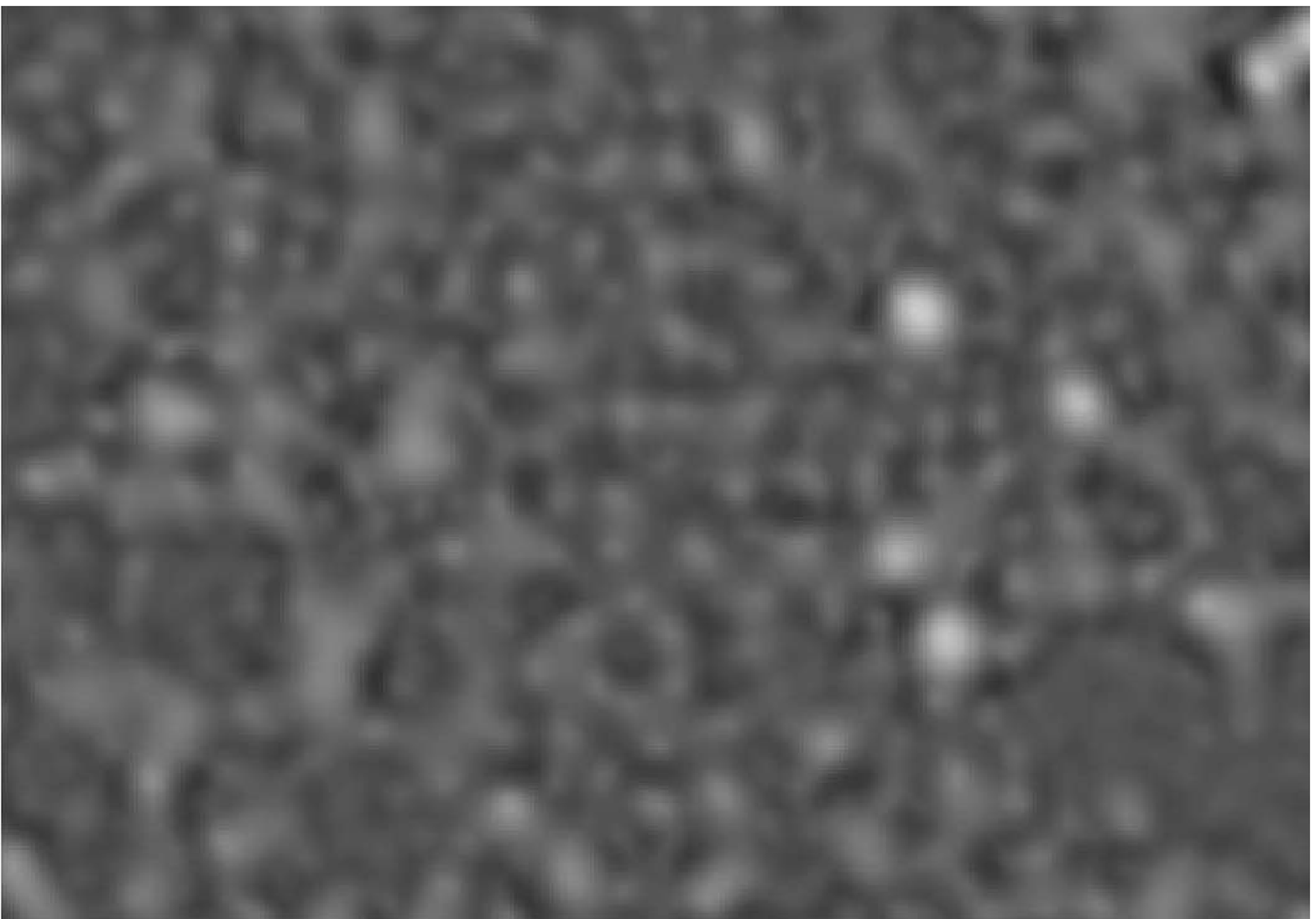}}
\subfloat[]{\includegraphics[width=.25\textwidth]{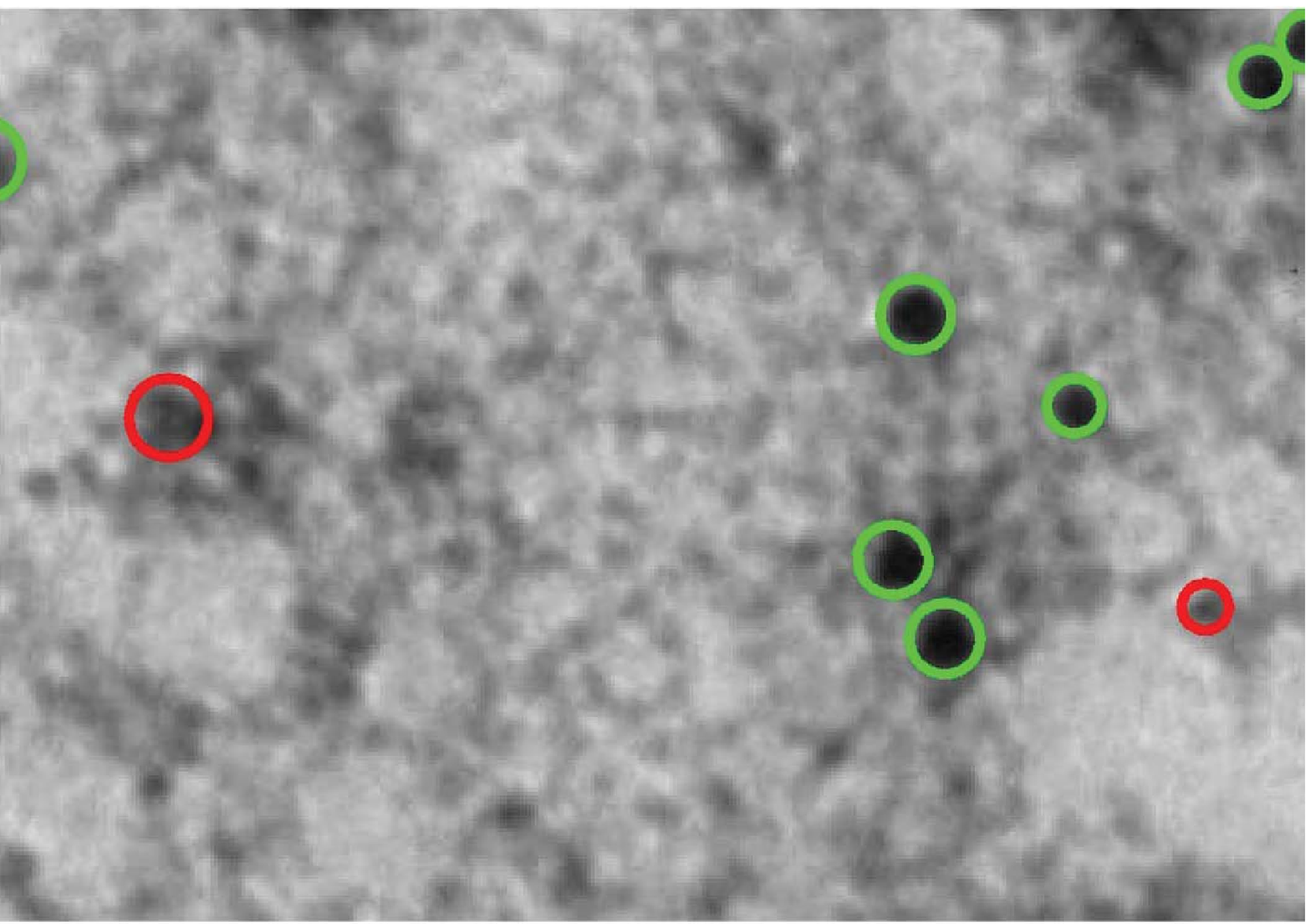}}

\subfloat[]{\includegraphics[width=.25\textwidth]{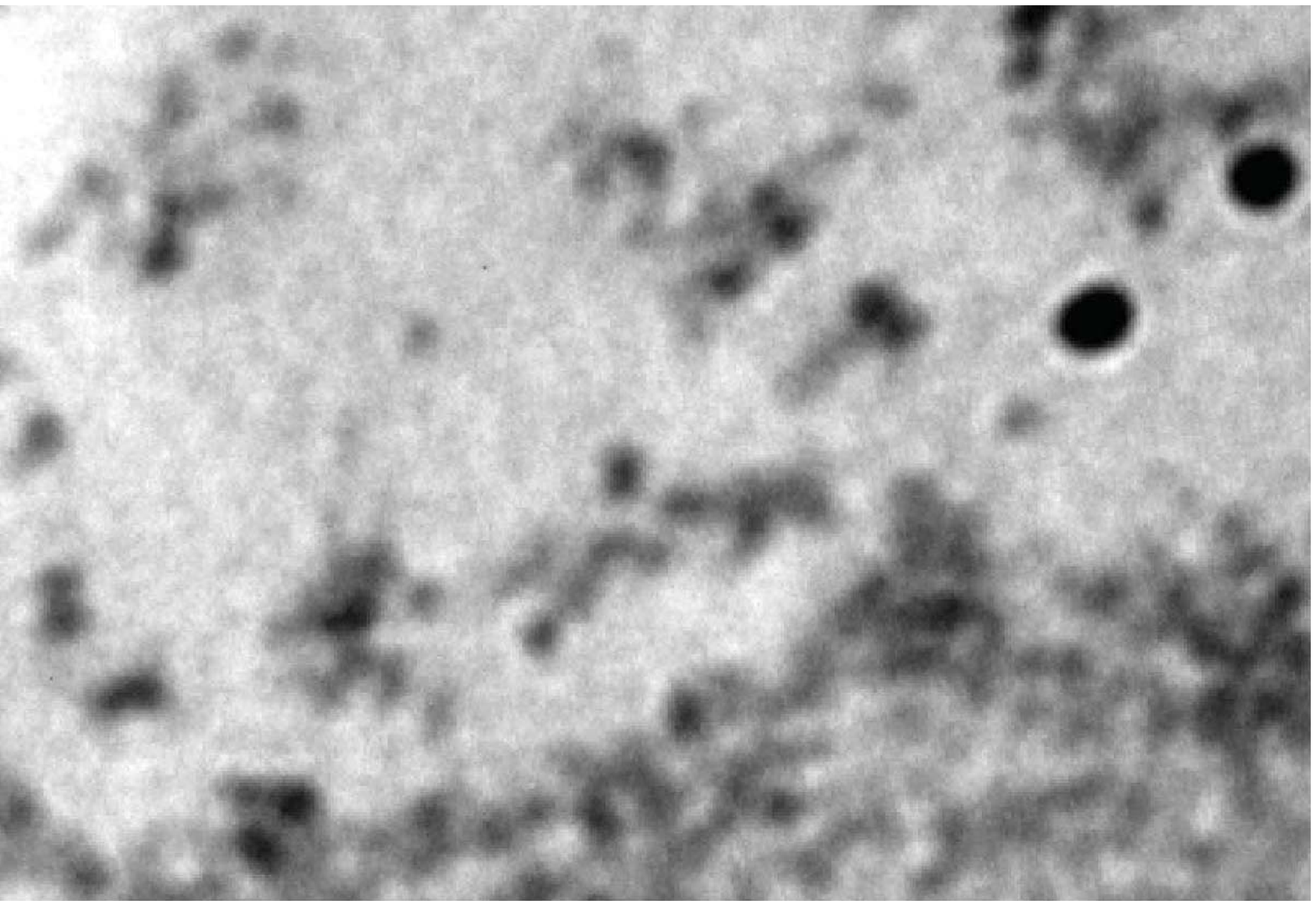}}
\subfloat[]{\includegraphics[width=.25\textwidth]{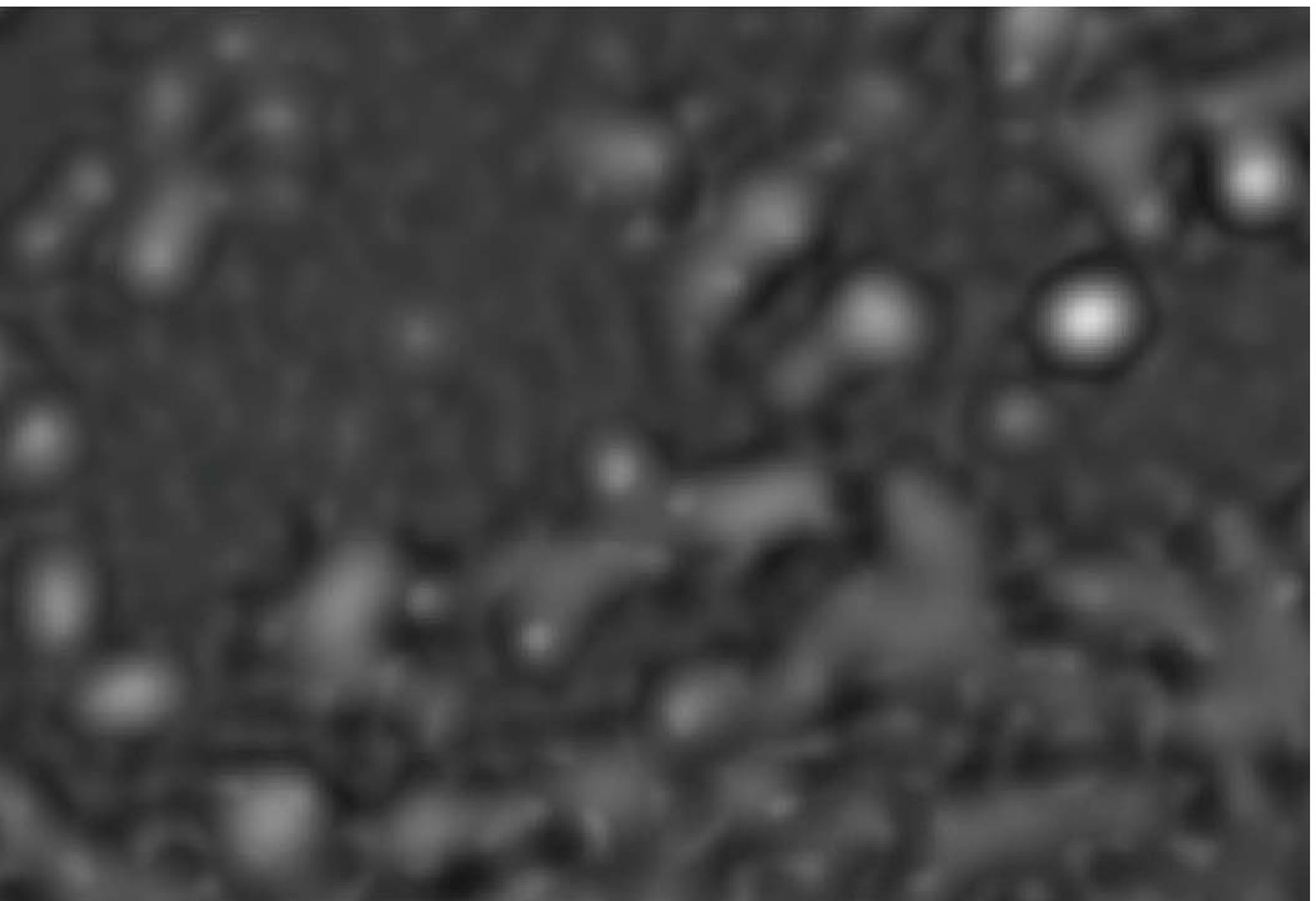}}
\subfloat[]{\includegraphics[width=.25\textwidth]{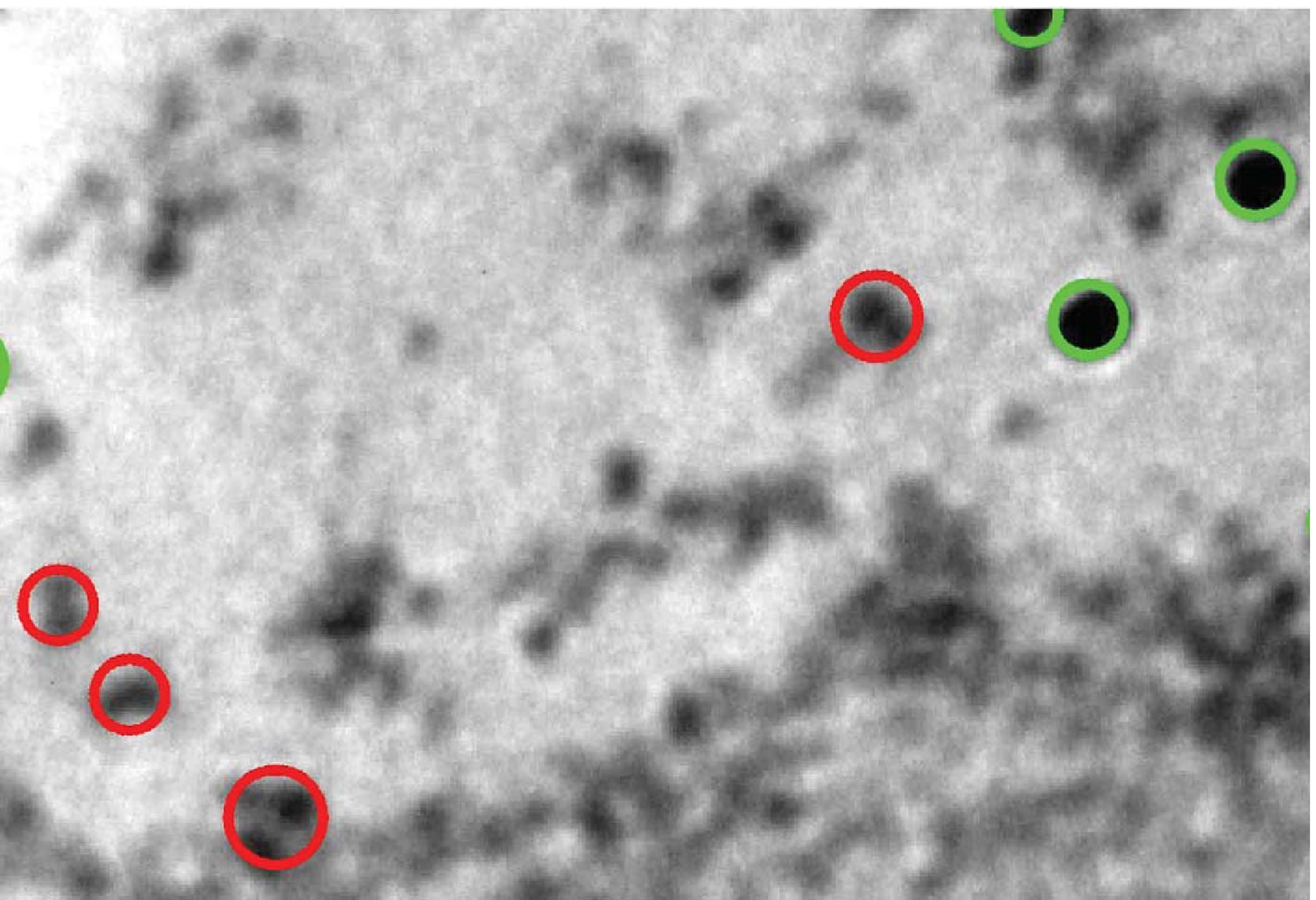}}

\subfloat[]{\includegraphics[width=0.25\textwidth]{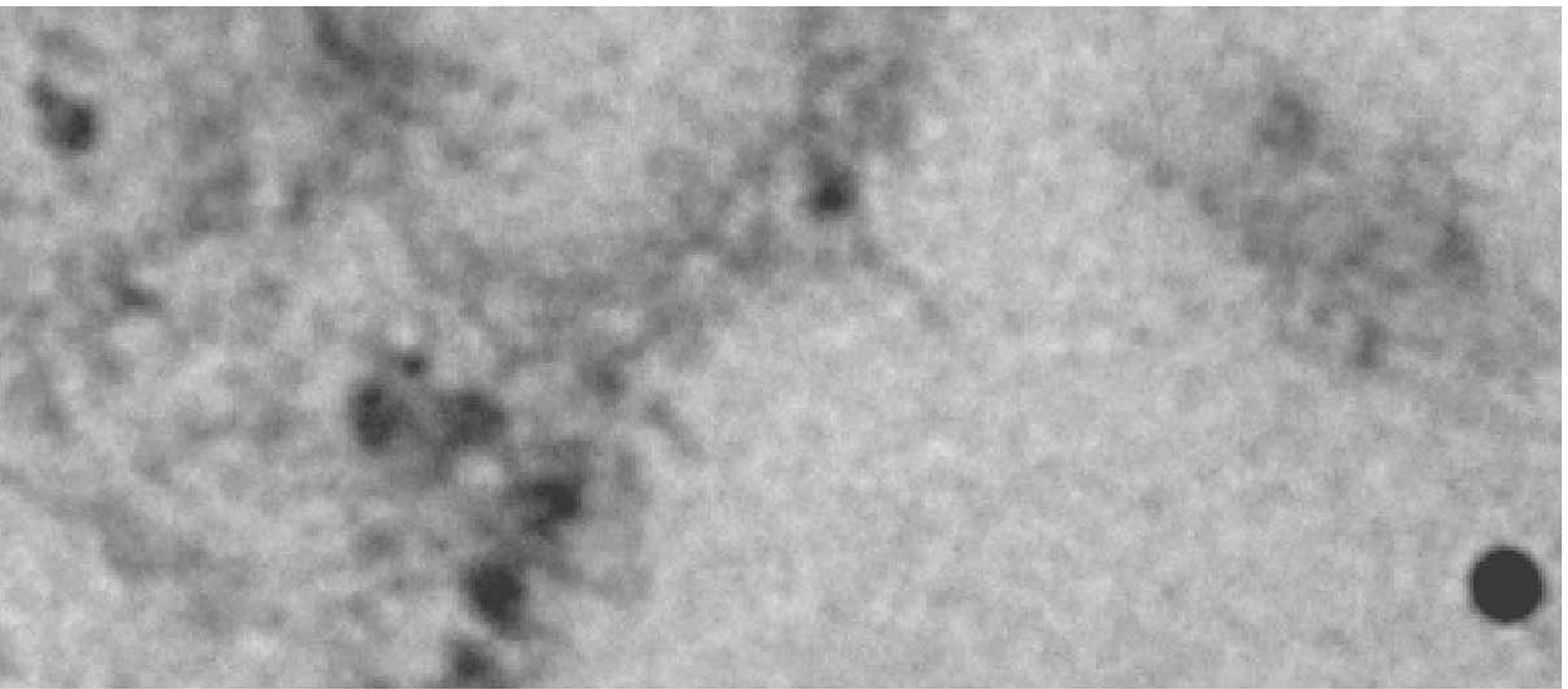}}
\subfloat[]{\includegraphics[width=0.25\textwidth]{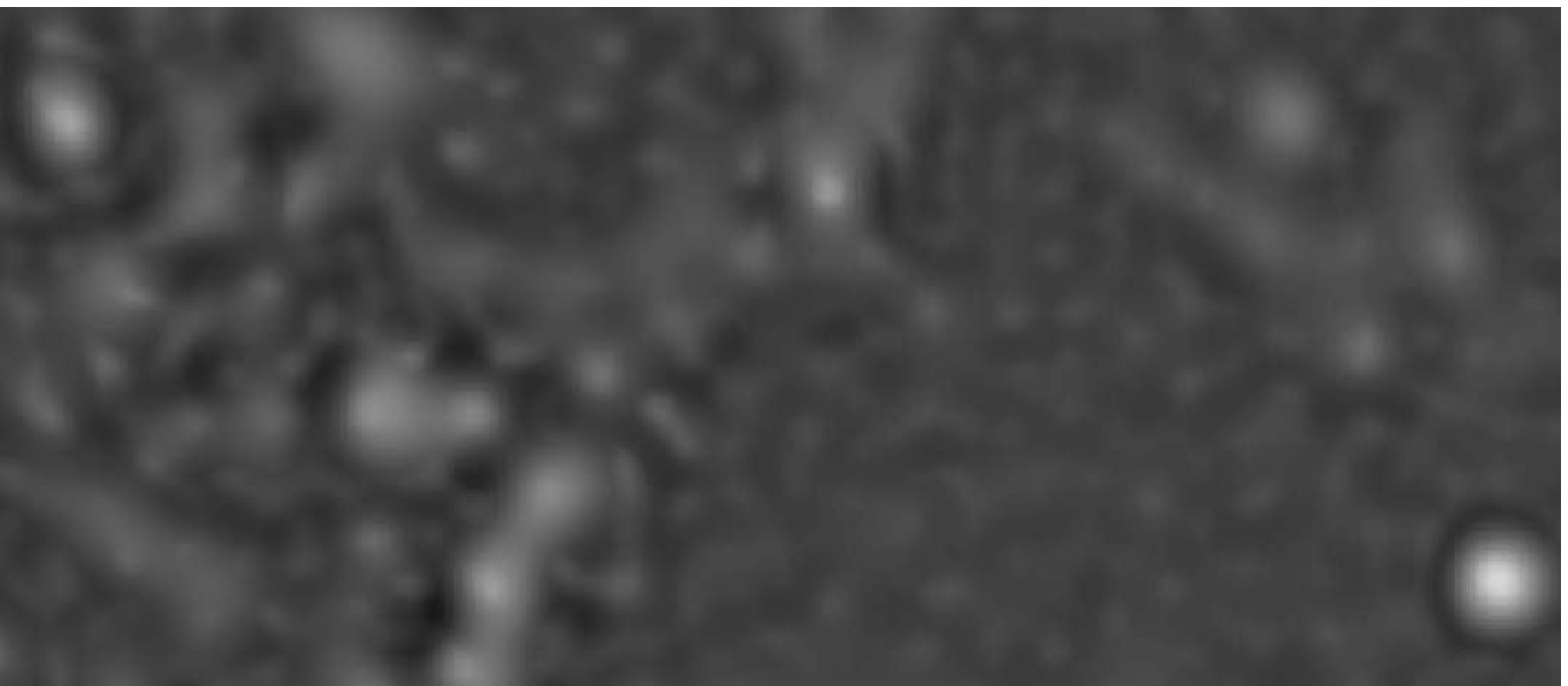}}
\subfloat[]{\includegraphics[width=0.25\textwidth]{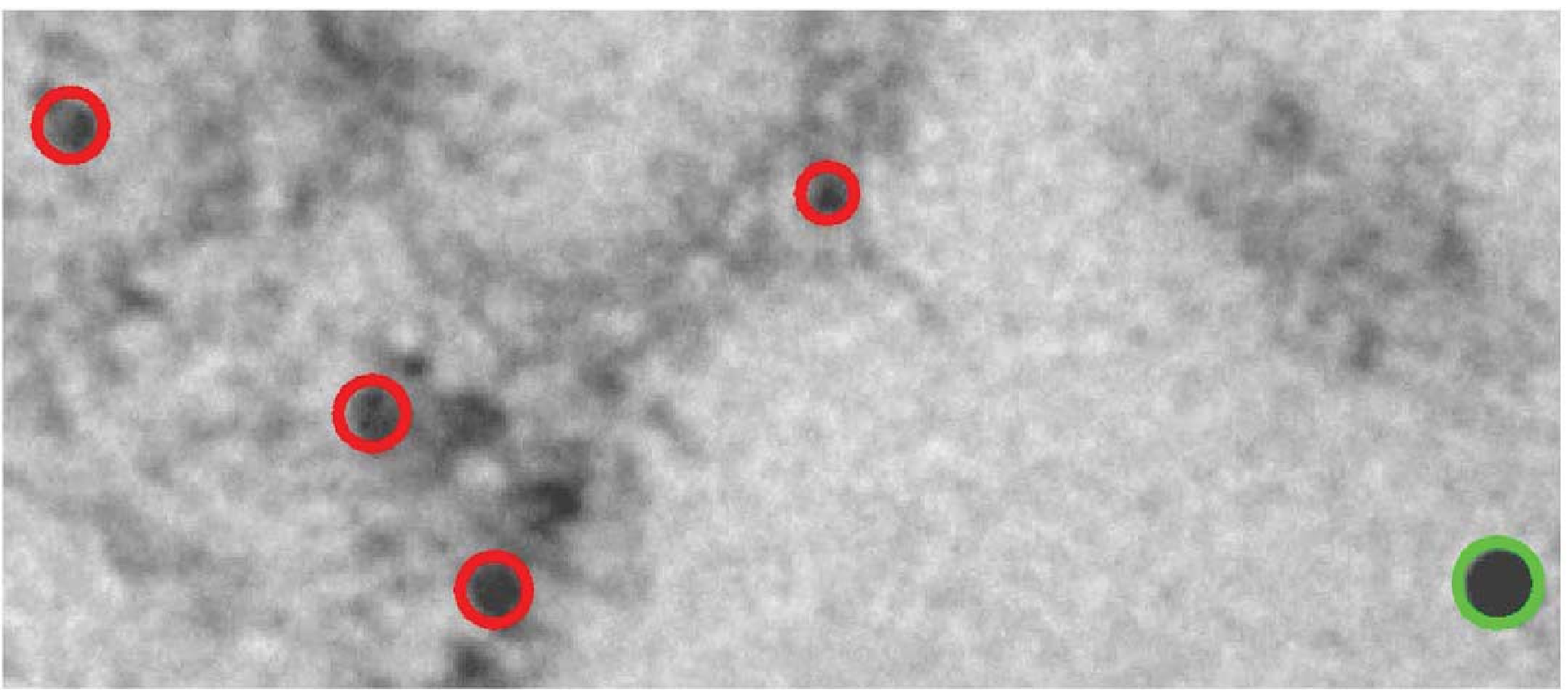}}
\caption{Immunogold particle detection based on the LoG filter. From top to bottom we have different detections from images with magnifications \db{15000} (\db{db1}), \db{20000} (\db{db2}), \db{30000} (\db{db3}) and \db{50000} (\db{db4}), respectively. For each row, on the left we have the original image followed in the center by the LoG response and finally on the right, the detections overlaid in the original image. In these samples we also show some false positive detections (in red) illustrating the difficulties of using the LoG filter alone (better viewed on color). Different sizes of the immunogold particles are related to the cropped region of the original image.}
\label{fig:logdetection}
\end{figure*}

It is known that the $\triangledown^2 L$ has a strong positive response for dark blobs of size $t$ and strong negative responses for bright blobs of similar size. However, and since possible variations on the immunogold particle structure can occur during the acquisition of the digital image of the biological sample (\fref{fig:dataset}), we have applied the \ac{LoG} filter over a range of possible scale responses. Given the expected range of the immunogold particles radius, $r$, we filtered the images over the range $t = \{r_i/1.5 \mid \forall r_i \in \{r-\delta,\ldots,r+\delta\} \}$. Setting $\delta = 1$ provided the best results for our experiments~\cite{RGamelasSousaICIAR2015}. We then perform the detection of immunogold particles by detecting local maxima in both spatial and scale dimensions over all \ac{LoG} responses (\fref{fig:logdetection}---center) in the input image (\fref{fig:logdetection}---left). The detected maxima enabled us to estimate the position of immunogold particles (\fref{fig:logdetection}---right).
A sample of the \ac{LoG} performance is depicted in \fref{fig:logdetection} for all datasets of this work: \db{db1} for the dataset of magnification \db{15000}, \db{db2} for the dataset of magnification of \db{20000}, \db{db3} for the dataset of magnification of \db{30000} and finally, \db{db4} for the dataset of magnification of \db{50000}.
Among the \ac{LoG} filter response and the respective detections, the erroneous detections provided by the \ac{LoG} are also visible in \fref{fig:logdetection}.

It can be claimed that other approaches can lead to similar (or even better results) than \ac{LoG}. Such approaches encompass the Circular Hough Transform~\cite{luo1995,pei1995} or variants of the Wavelets Transform~\cite{olivo2002extraction}. One variant of the wavelets transform is publicly available in the bioimaging software identified as \acf{SD} and which we have already mentioned above. The Circular Hough Transform is a particular case of the Hough Transform specially tailored for the detection of circular objects. In a nutshell, this method consists on fitting a circle for a given interest point. Applying these approaches to the immunogold particle detection problem rendered however a significant reduction of performance in comparison to the \ac{LoG}. For these reasons, and since this sole approach can induce significant detection errors, we have focused on coupling a robust \ac{ML} technique to the \ac{LoG} filter.

\subsection{Immunogold Recognition using Stacked Denoising Autoencoders}
In order to speed-up the detection and recognition processes, the first step of our framework is performed by the \ac{LoG} filter. We then processed the detections provided by the LoG, by employing a deep learning algorithm.

\begin{figure*}[]
\centering
\subfloat[]{\includegraphics[width=.20\textwidth]{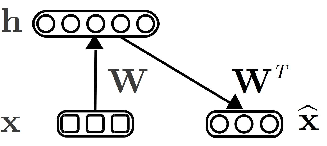}}
\hspace{1cm}
\subfloat[]{\includegraphics[width=.25\textwidth]{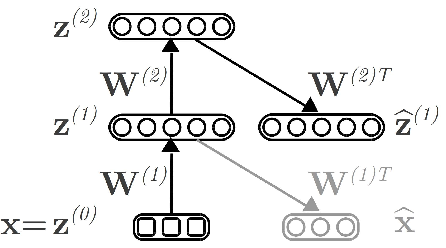}}
\hspace{1cm}
\subfloat[]{\includegraphics[width=.20\textwidth]{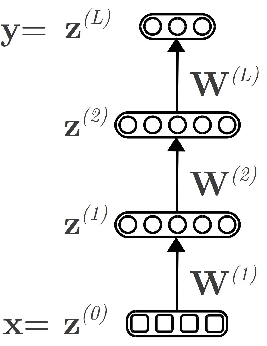}}
\caption{(a) An auto-encoder, (b) Pre-training of hidden layers of a deep network using auto-encoders. (c) A complete deep network with two hidden layers and an output layer. (Based on~\cite{larochelle07}).}
\label{fig:sda}
\end{figure*}

An \ac{AE} is, in the simplest form, a \ac{NN} with one hidden and one output layer that is designed to reconstruct its own input. As exemplified in Fig.~\ref{fig:sda}(a), it is subject to two restrictions: the weight matrix of the output layer is the transposed of the weight matrix of the hidden layer and the number of output neurons is equal to the number of inputs~\cite{Telmo2013}.

The values of the hidden layer neurons, called the encoding, are obtained via Equation~\ref{eq:ae_enc}, where $\mathbf{x}$ is the input vector, $s$ denotes the sigmoid function, $\mathbf{b}$ is the vector of hidden neuron biases, and $\mathbf{W}$ is the matrix of hidden weights. The values of the output neurons, called the decoding, are computed as in Equation~\ref{eq:ae_dec}, where $\mathbf{c}$ is the vector of output neuron biases. One advantage of the \acp{AE} is their ability to capture relevant information of the underlying distribution of the samples~\cite{rifai2011}.
\begin{equation}\label{eq:ae_enc}
    \textbf{h}(\textbf{x}) =s\left(\textbf{b}+\textbf{W}\textbf{x}\right)
\end{equation}
\begin{equation}\label{eq:ae_dec}
    \hat{\textbf{x}}(\textbf{h}(\textbf{x})) =s\left(\textbf{c}+\textbf{W}^{T}\textbf{h}(\textbf{x})\right)
\end{equation}

Though \acp{AE} attain sparse features with good generalization capabilities, it is still required to have a model that captures good representations of the data. To achieve this goal the \ac{DAE} has been presented as a robust generalization of the \ac{AE}. \acp{DAE} models are still designed to rebuild the original data but now from input data corrupted with noise. In practice, a small percentage of random components of $\textbf{x}$ are set to zero~\cite{Bengio2008}. Reconstruction of the input data is achieved by minimizing the reconstruction loss while allowing the \acp{DAE} to avoid a direct copy of the data~\cite{vincent2010}. These models had their robustness augmented by stacking (denoising) autoencoders (Fig.~\ref{fig:sda}(b)). \ac{SDA} gives the model the advantage of learning hierarchical features with low-level features represented at lower layers and higher-level features represented at upper layers~\cite[Section 3]{bengio2012deep,vincent2010}. One limitation of these ``deep’’ models is concerned with the number of layers that one has to train and the total amount of time that it would take.

The breakthrough for training these models was achieved by conducting a layer-by-layer unsupervised training to find the proper initialization weights. This scheme allows the learning model to escape from local minima that is so common when training such networks~\cite{Bengio2008}. This unsupervised training approach of the \acp{SDA} is usually referred to as pre-training (Fig.~\ref{fig:sda}(b)). To perform a classification task, we add to the top of the network a logistic layer (Fig.~\ref{fig:sda}(c)) and the entire network is then ``fine-tuned'' in order to minimize some classification loss function~\cite{Telmo2013,bengio2012deep}.

\begin{figure}[!ht]
\centering
\includegraphics[width=.48\textwidth]{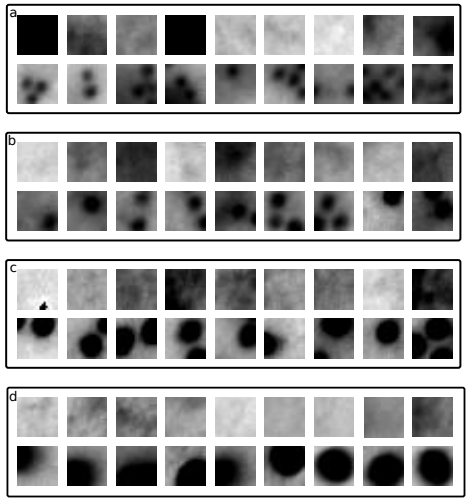}
\caption{Different types of patches for all databases: (a) \db{db1}, (b) \db{db2}, (c) \db{db3} and (d) \db{db4}. In each, the top row shows background, noise and artifacts and the bottom row shows immunogold particles.}
\label{fig:dbsae}
\end{figure}

\acp{SDA} are a very promising \ac{ML} tool for the recognition of the circular shaped immunogold particles. A representative sample of the images that we used to train the \ac{SDA} is depicted in \fref{fig:dbsae}.

Given the particular features of the immunogold particles, it is expected that an \ac{SDA} captures relevant representations from these samples and easily discriminates the immunogold particles from the remaining artifacts or cellular structures. A far more complex scenario occurs when multiple immunogold particles are present in the same patch (i.e, a small window), (see \fref{fig:dbsae}), especially in low magnifications such as in the \db{15000} magnification. In this case an \ac{SDA} has to be able to deal with a variable number of particles in the same patch.
Thereby, an \ac{SDA} is trained to recognize particles belonging to a given magnification. To strengthen the robustness of this model to noise and other artifacts we train it by corrupting the training samples with noise. Once the best model is obtained it will consider the estimates provided by the \ac{LoG}. The \ac{SDA} thus considers a patch of the image centered in the location given by \ac{LoG} to output the final prediction.

\section{Transfer Learning for the Recognition of Immunogold Particles}
\label{sec:TL}
\ac{TL} aims to transfer a learning model that was acquired in one problem, the source problem, onto another problem, the target problem, dispensing with the bottom-up construction of the target model~\cite{RGamelasSousaIWANN2015}. With this, \ac{TL} algorithms aim to overcome one of the major drawbacks of \ac{ML}: limited labelled data. There is another advantage of using \ac{TL} algorithms: reduction of development and computational times for learning (training) models for new (target) problems. TL algorithms embedded in \acp{DNN} have already shown interesting features such as better generalization capabilities and reduced training times as shown in other works~\cite{Chetak2014} (and references therein).\footnote{Check also the website: \url{http://www.deepnets.ineb.up.pt}}

As it was clear in the previous Section, our models' performance is also bounded by the quality of the dataset (and its annotations). For instance, on datasets with low-magnifications (i.e., \db{15000}) we have noise and complex structures. Ergo, we obtain a feeble performance or high variances for the recognition of particles on such low-magnifications mostly because of those artifacts. For this scenario the advantage on using \ac{TL} should be obvious: to transfer a model that was obtained on a higher magnification to a dataset of a lower magnification. Consequently, \ac{TL} presents itself as an appealing alternative.
Under this broad topic there has been a diverse set of methods being proposed: Never-ending learning~\cite{thrun1996learning}, covariate shift (ref. [20] in~\cite{pan2010survey}), weighting instance (ref. [6] in~\cite{pan2010survey}) and unsupervised TL (ref. [26] in~\cite{pan2010survey,mesnil2012unsupervised}) are just a small set of methodologies presented in the literature. See~\cite{pan2010survey} for a broader analysis. \acp{DNN} have raised some interesting questions within this research topic. Concretely, how can we devise \acp{DNN} for \ac{TL} and what are the best strategies for these algorithms~\cite{kandaswamy2014,RGamelasSousaIWANN2015}. On the other hand, Patricia \etal proposed in~\cite{patricia2014} a single Support Vector Machine formulation to solve the different types of \ac{TL} settings.
Beker \etal in~\cite{becker2014} worked on the application of TL to Microscopy Imaging and presented a covariate shift approach for the recognition of Cellular Structures using a set of week classifiers. However, the experimentations and adaptability of \ac{TL} has been limited to the analysis of Mitochondria.

In this Section we will describe our framework under \ac{TL} consisting on the following points: (a) An \ac{SDA} is first trained to distinguish immunogold particles from cluttered background; (b) This model is then reused on another problem (with a different magnification); and, (c) Models are then assessed for the detection task.

Low-magnifications (e.g., magnification of \db{15000}) are usually used by a life-scientist for identifying regions of interest in \ac{EM}. Even though these acquisitions are afterwards discarded to perform a more in-depth analysis, they contain rich information that can expedite many of the quantifications that have to be conducted for a given experiment.
By directly exploring images obtained at lower magnifications we can attain reduced experimentation costs, image acquisition and labelling times.
However, an automatic learning model is prone to misclassify a large set of patches due to the noise and artifacts present in the dataset (we will see some of these effects shortly). To overcome this undesirable outcome, we can build a learning model using a dataset with a high magnification as source problem (e.g., \db{50000}, containing well defined immunogold particles and without clusters of immunogold particles), and transfer it to a target problem of a dataset with a low magnification (e.g., \db{15000}).
Putting it simply, layer weights for the target model are initialized with the values of the source model. Learning (of the target problem) can be conducted in several ways: (a) by fixing the layer weights of the network or by letting them readjust through the minimization of the reconstruction error (pre-training); (b) by fixing or letting the network relearn and thus letting it to readjust the decision function (fine-tune); or (c) a mixture of both~\cite{Chetak2014,Chetak2014b,Telmo2014,bengio2012deep,theano2012} (and references therein) --- see \algoref{algo:tl} for a brief presentation.

For the purpose of this work, we considered the (b) approach above described. In our experimental work we have analyzed the impact of the reusability of the different hidden layers. In fact, and for the immunogold particle recognition, one may argue that the major changes will occur at a feature level thanks to the changes of the magnification of the immunogold particles. Looking carefully at \fref{fig:dataset} we can see that this seems to be the case. Therefore, the recognition performance of the immunogold particles through \ac{TL} would have significant impact by relearning only the first layers of the network. The analysis of the effects of relearning specific parts of the network architecture will be addressed in more detail in \sref{sec:results}.

\begin{algorithm*}
  \caption{Pseudocode for training a \ac{TL} model using an \ac{SDA} for the recognition of immunogold particles.}
  \label{algo:tl}
  \begin{algorithmic}[1]
    \State \textbf{Input}: ${\cal M_S}$, an \ac{SDA} model trained on a source problem and $h_{TL}$, the hidden layers that one wishes to reuse.
    \State \textbf{Output}: ${\cal M_T}$, an \ac{SDA} model obtained through TL for the target problem.

    \SetKwFunction{FineTuneNetwork}{FineTuneNetwork}

    \State ${\cal M_T} \leftarrow {\cal M_S}$ \tcc{Copy the source model}
    \State Let ${\cal H}$ be the set of hidden layers of ${\cal M_S}$ (or ${\cal M_T}$) and $h = h_{TL} \setminus {\cal H}$\;
    \State Discard the hidden-layers $h$ of the \ac{SDA} target model, ${\cal M_T}$\;
    \State ${\cal M_T} \leftarrow$ \FineTuneNetwork(${\cal M_T}$,$h$)\;
  \end{algorithmic}
\end{algorithm*}

\section{Experimental Study for Immunogold Detection and Recognition}
\label{sec:results}
In this Section we will thoroughly describe the experimental work that we have conducted as well as a detailed discussion on the results that were obtained. The code is publicly available\footnote{Available at \url{http://www.deepnets.ineb.up.pt}}. Our study is split into two halves: first, the detection by coupling the \ac{SDA} with the \ac{LoG} filter and second, the recognition of immunogold particles.

\paragraph{Immunogold Particles Dataset:}
This novel dataset results from a study of the composition of cell wall ingrowths of maize endosperm transfer cells via the detection and quantification of immunogold particles. As previously stated, images with lower magnification allow a wider field of view but the detection of the immunogold particles is much more difficult than with higher magnifications. The purpose was to develop an automated system that could be able to detect particles in lower magnification images with a comparable performance to that obtained with higher magnification images. The dataset contains 100 images with 4008 pixels wide and 2670 pixels tall. All images were acquired using a \acf{TEM} JEOL JEM 1400 with a GATAN Orius SC10000A2 CCD. These images were recorded in four different magnifications: \db{15000} (scale of $1\mu m$ to the size of the image), \db{20000} ($0.5\mu m$), \db{30000} ($0.5\mu m$) and \db{50000} ($200 nm$) from different biological samples (see \fref{fig:dataset}) whereas a manual annotation was conducted with the plugin `manual counting' provided by Icy software~\cite{de2012icy}. Each magnification corresponds to a single dataset (see Table 1). The datasets are publicly available on our website.

\paragraph{Performance Measures:}
The detection performance of the \ac{LoG} filter was conducted by assigning a ground-truth to a given detection if the Euclidean distance between the ground-truth and the detected point positions is below the size of the particle radius $r$. Of course, we ensure that there is a one-to-one mapping between detection and ground-truth. To measure the performance of the automatic detections we used the following metrics:
\begin{itemize}
\item True Positive (TP): number of correctly detected immunogold particles;
\item False Positive (FP): number of falsely detected immunogold particles;
\item False Negative (FN): number of immunogold particles which were not detected.
\end{itemize}
The performance of our method was plotted in a Precision-Recall curve~\cite{davis2006} as follows:
\begin{equation}
   \text{Precision}  = \frac{TP}{TP+FP}, \qquad  \text{Recall} = \frac{TP}{TP+FN}
\label{eq:prec_rec}
\end{equation}
We will also summarize the performance results of our methods by using the F-measure which combines both precision and recall. This measure is given as follows:
\begin{equation}
   \text{F-measure}  = 2 \cdot \frac{\text{Precision} \cdot \text{Recall}}{\text{Precision}+\text{Recall}}
\label{eq:prec_rec1}
\end{equation}

\subsection{Detection and Recognition of Immunogold Particles}\label{sec:expLOG}
Our experimental study starts by first constructing an \ac{SDA} model. We extract patches with $20\times 20$ pixels from 60\% of our dataset (60 images, 15 samples per magnification) half of them containing all the immunogold particles and the other half containing background, cellular structures or other artifacts --- see \tref{tab:DBSIZE}.
A patch could contain more than one particle and/or portions of several other particles (see \fref{fig:dbsae}). Finally, patches are labelled as containing at least one immunogold particle if the Euclidean distance between the patch position (on the image) and the position of the annotation is below the size of the patch.
Pixel values of the patches were normalized to be within $[0,1]$. To find the best parametrization of the \ac{SDA} model we have performed a grid search on the pre-training learning rate $[0.01, 0.001]$ and fine-tune learning rate $[0.1, 0.01]$.
We fixed the number of neurons to $1000$ units per layer and the number of hidden layers was also fixed to $3$.
The grid search was conducted by carrying out a three-fold cross-validation in the training set and performance assessed in the validation set. The corruption level was set to 10\% across all hidden layers.
Pre-train and fine-tune epochs were set to 1000 and 3000, respectively, and all models were trained by batches. For the \db{db1} we used $1000$ patches per batch and for the \db{db2}, \db{db3} and \db{db4} we used $100$ patches per batch.
For the execution of our \ac{TL} experiments, we had to resize by half the images of the datasets \db{db3} and \db{db4} to afterwards use them as source problem (more details for this results will be described thoroughly in \sref{sec:expTL}). Note, however, that patches used to train our models had the same dimension, i.e., $20 \times 20$ pixels.
Our models were developed and tested using the Theano framework~\cite{theano2012,theano2010} with simulations being executed on a GTX 980 GPU on a i7-5930K CPU with 16Gb of RAM.
Test results of the \ac{SDA} are depicted in \tref{tab:sda_results}.

\begin{table}[!htp]
  \centering
  \caption{Size and magnification of each dataset used to train our models. Datasets are balanced being composed by the same number of immunogold particles (objects) and cellular structures or other artifacts (background).}
  \label{tab:DBSIZE}
  \begin{tabular}{c|cccc}
    \hline\hline
                 & \db{db1} &  \db{db2} & \db{db3} & \db{db4}\\
    \hline
    Number of instances & 16556    &  10766    & 5060     &  1644\\
    Magnification & 15000  & 20000  & 30000 & 50000 \\
    \hline\hline
  \end{tabular}
\end{table}

\begin{table}[!htp]
  \centering
  \caption{Baseline results for the recognition of immunogold particles with \acp{SDA}. We also include the pre-training and fine-tuning times for training this model (average time per repetition).}
  \label{tab:sda_results}
  \begin{tabular}{p{1cm}cp{2cm}p{2cm}}
    \hline \hline
    Dataset & Accuracy & Pre-training Time (sec.) & Fine-Tuning Time (sec.) \\
    \hline
    \db{db1} &  $79.1 \pm \p5.6$ & $ 225.1  \pm 44.5$ & $\p30.0  \pm 49.3$\\
    \db{db2} &  $96.0 \pm \p1.4$ & $ 245.3  \pm 51.0$ & $ 367.8  \pm 81.6$\\
    \db{db3} &  $87.1 \pm  12.6$ & $ 135.7  \pm 18.6$ & $183.5  \pm  78.1$\\
    \db{db4} & $91.5 \pm \p5.2$ & $\p33.7  \pm 10.0$ & $\p54.3 \pm \p8.0$\\
    \hline\hline
  \end{tabular}
\end{table}

Once we obtained the best \ac{SDA} model, we proceeded with the development of the \ac{LoG} filter. \ac{LoG} is solely controlled by the scale parameter which we have set to vary between known values of the immunogold particles size: $[3,\ldots, 13]$ pixels. For experimental swiftness, and since some particle radius could not make sense for a given magnification, we trimmed the search space for an ``optimum’’ radius value. Thus, the radius range was set to $\{3,4,5\}$, $\{3,5,7,9\}$, $\{5,7,9,11\}$ and $\{9,11,13\}$, pixels, for images of the datasets \db{db1} to \db{db4}, respectively.

\begin{figure*}[t]
\begin{center}
\subfloat{
  \label{fig:15000}
  \includegraphics[width=0.48\linewidth]{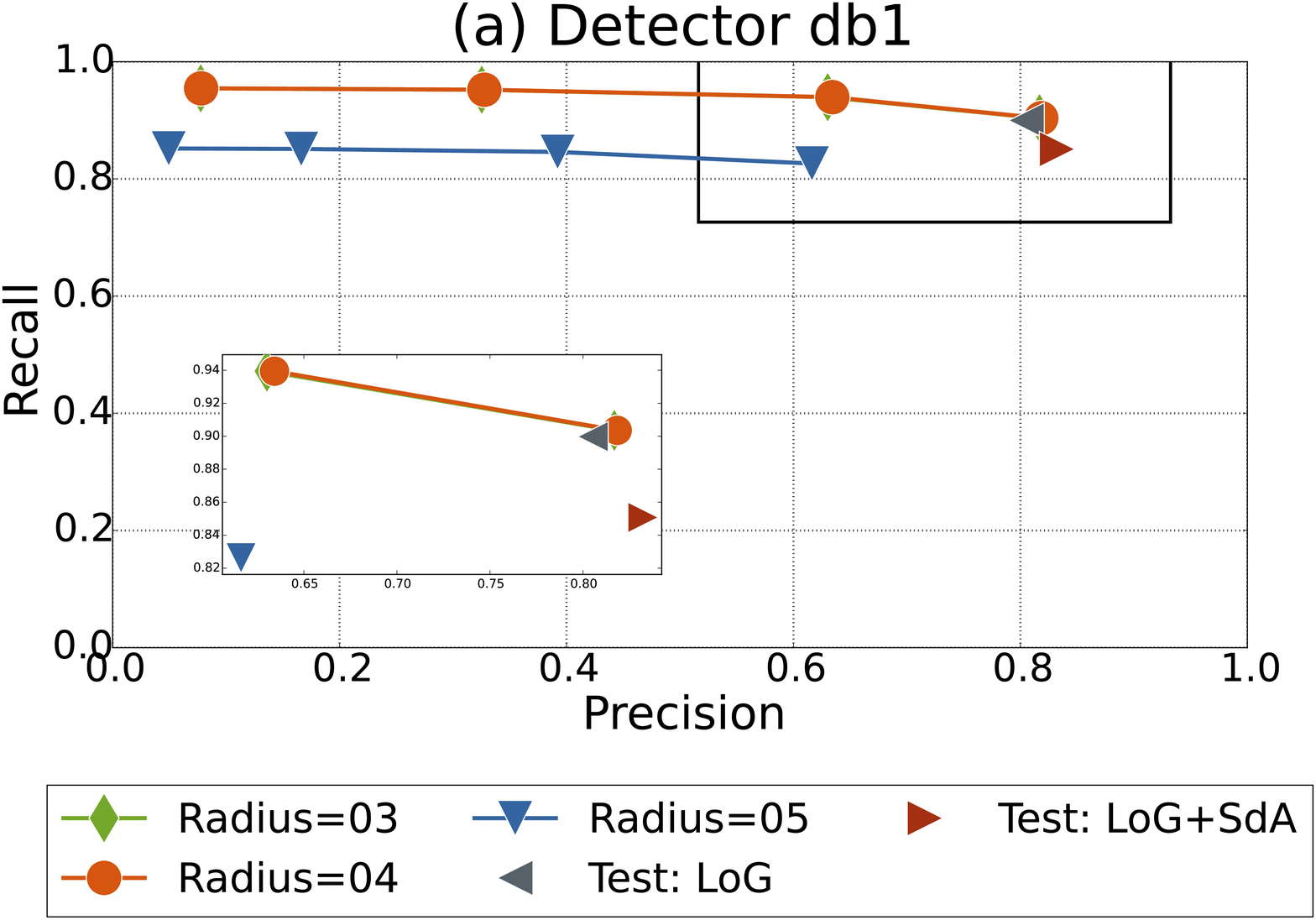}
}
\hfill
\subfloat{
  \label{fig:20000}
  \includegraphics[width=0.48\linewidth]{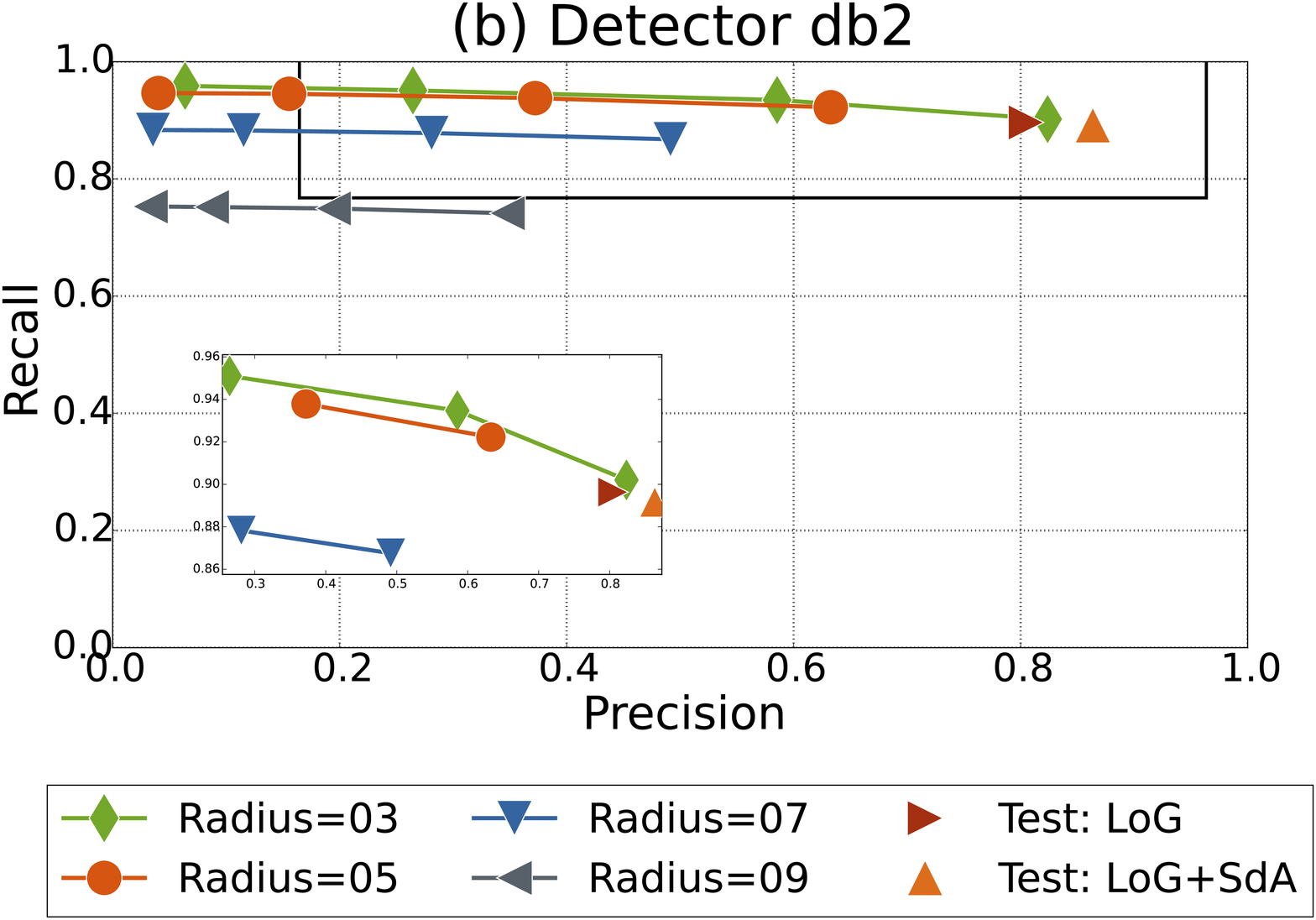}
}
\hfill
\subfloat{
  \label{fig:30000}
  \includegraphics[width=0.48\linewidth]{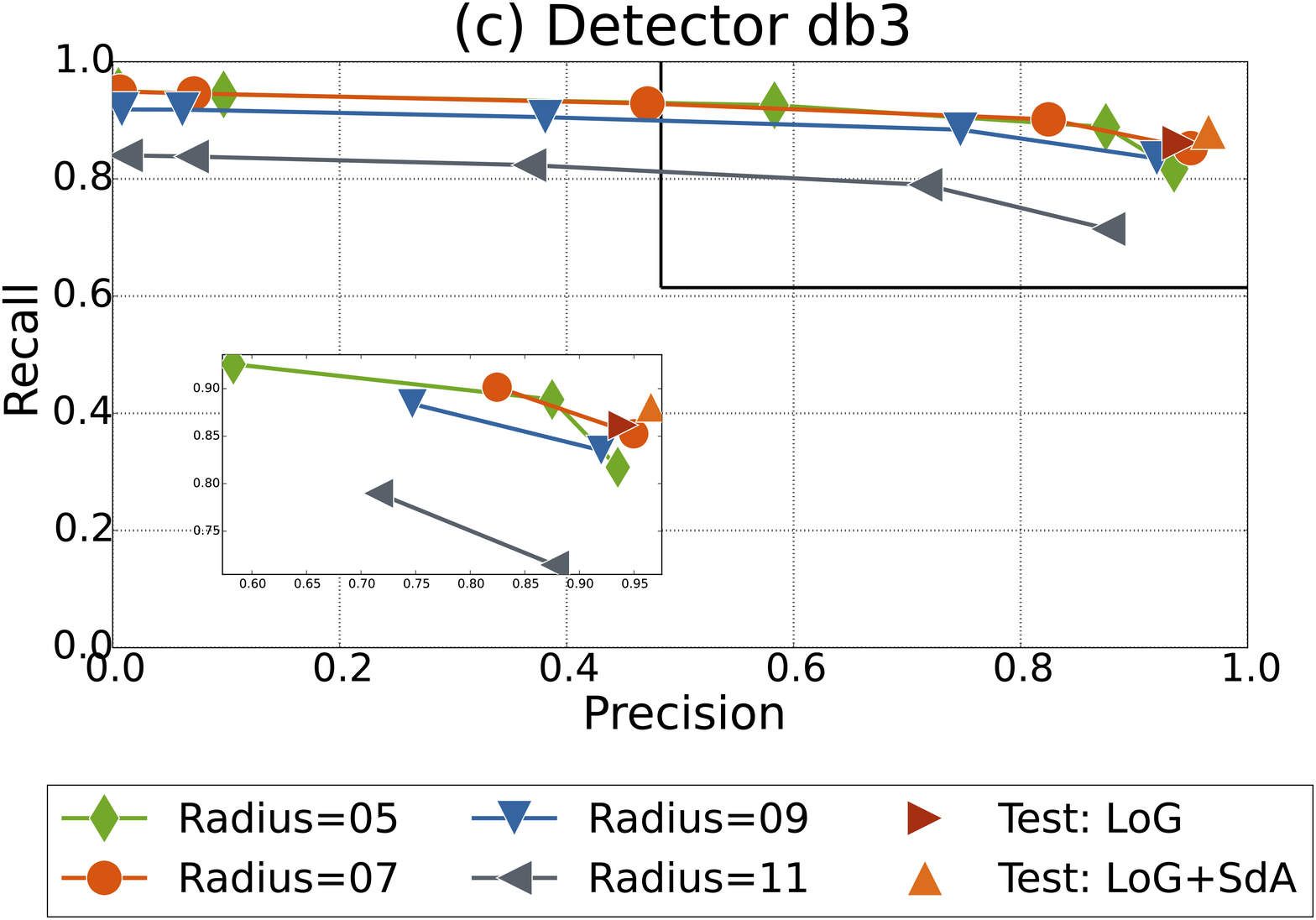}
}
\hfill
\subfloat{
  \label{fig:50000}
  \includegraphics[width=0.48\linewidth]{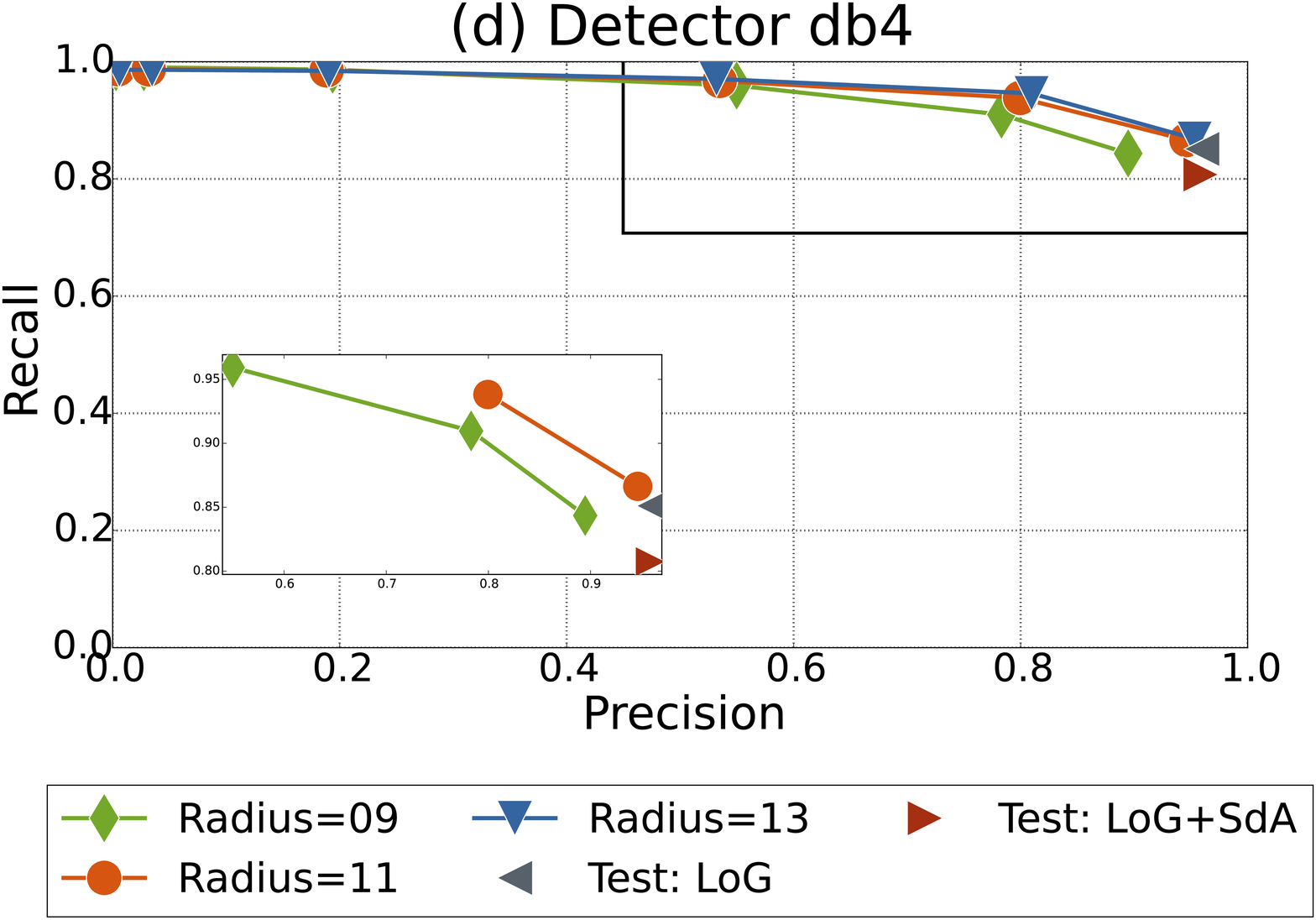}
}
\caption{Precision-Recall curves averaged over 20 repetitions for the \ac{LoG} on the validation set. Non-connected points correspond to the test results for \ac{LoG} and \ac{LoG}+\ac{SDA}. Upper right areas of interest in each graph is zoomed in larger rectangles for visualization purposes.}
\label{fig:log_sda_results}
\end{center}
\end{figure*}

We also measured the performance for each threshold applied to the filter response. This was done by ranging the values of the threshold between 5 and 55. The threshold set had to be adjusted for a given magnification: $\{10,15,20,25\}$ for \db{db1} and \db{db2} and, $\{5,\ldots,45\}$ with steps of 5 for \db{db3} and $\{5,\ldots,55\}$ also with steps of 5 for the \db{db4}.
To find the best parameterization we have performed a three-fold cross validation with the \emph{same} 60\% of the total number of images (60 images, 15 samples per magnification) as in with the \ac{SDA} for training our \ac{LoG} model and with the \emph{same} 40\% for evaluating its performance. In other words, the learning process was conducted independently from the \ac{LoG} training, but using the same images. The performance of the \ac{LoG} is represented by a Precision-Recall curve on the validation set~\cite{davis2006} and results are shown in \fref{fig:log_sda_results}.

This framework works as follows: Given a test image we first apply the \ac{LoG} filter to obtain the initial estimates of the immunogold particles position. Afterwards, we extract a patch with $20 \times 20$ pixels centered at the position of the \ac{LoG} detection. This patch is afterwards evaluated by the \ac{SDA} that we developed before. Finally, we assessed the variability of our methods' performance by repeating the experiment 20 times by shuffling the dataset.

For the magnification of \db{15000} --- see \fref{fig:15000}, we observed that the expected immunogold particle radius tested during the detection of immunogold particles attained the best results ($radius = 3$ and $4$). Comparing to the other radius tested, with $radius = 3$ (or $4$) it achieved a higher recall, which means a higher number of immunogold particles were detected. A similar analysis can be performed for the remaining figures stating the importance of the size of the radius.
Moreover, by having one single parameter this method is intuitive and easy to use though prone to some error as it is expressed by the low precision rates. Another important result is that when using higher magnifications, \db{30000} and \db{50000}, we can obtain very good results only with the \ac{LoG} filter. This behavior is related to the quality of the images which contain little noise and artifacts. Moreover, these datasets also contain immunogold particles far apart from each other, making them easy to detect. Unfortunately, when working with such high magnifications (e.g., \db{50000}) the field-of-view is reduced significantly. Thus, working with such high magnifications may be useful only after determining the regions of interest.

\subsection{Evaluation of TL for the Recognition of Immunogold Particles}\label{sec:expTL}
As it was mentioned, one advantage of using \ac{TL} algorithms is that it eliminates the need to redesign learning models for a distinct, but related, problem. In our case, since we are using several magnifications of \ac{EM} images, we explore different \ac{TL} settings for the recognition of immunogold particles of Maize cells.

The experimental procedure to train the baseline models was conducted as described above in \sref{sec:expLOG}. We then proceeded with the \ac{TL} approach. Images of the dataset \db{db4} were resized to half (keeping the aspect ratio), but maintaining the size of the patches ($20\times 20$). This led to immunogold particles with a radius around 7 pixels (recall that the radius size of immunogold particles on the dataset \db{db1}, was 4 pixels). If we do not resize the images of the source problem we do obtain some accuracy improvements but lower than the ones when resizing the source problem.
\ac{TL} can be conducted by reusing the first layer of the source model and by fine-tuning only the remaining layers to the target model (coded [011]); another possibility is reusing the first and second layers and fine-tuning the last layer ([001]). By reusing the full source network ([000]) the model would not suffer changes, and thus would have not been fine-tuned to the target problem. Results for all variations of the \ac{TL} settings are presented in \tref{tab:TLdb4}.

\begin{table*}[!htp]
  \centering
  \caption{Results of the application of \ac{TL} to the recognition of immunogold particles. The baseline model was trained in a standard \ac{ML} way on the dataset with magnification of \db{15000} (target problem). A model trained for the dataset with magnification of \db{50000} (source problem, immunogold particles were resized) was obtained and reused on the target problem. Overall, all \ac{TL} approaches achieved an improvement of more than 10\%. Each layer-wise \ac{TL} strategy is illustrated in the column \ac{TL} setting (see main text). Results were averaged over 20 repetitions.}
  \label{tab:TLdb4}
  \begin{tabular}{p{1.5cm}p{1.3cm}p{1.cm}>{\centering\arraybackslash}p{2.5cm}p{2cm}p{3cm}}
    \hline\hline
    Method  & Source (resized)          & Target       & Reusability and Fine-Tuning (TL) Setting & Accuracy ($\pm$~std. dev) & Pre-Training + Fine-Tuning Time (sec.)  \\ \hline
    Baseline&    -     & \db{db1}   &  -           & $79.1 \pm 5.6$ & $255.1 \pm 83.0$ \\
    \hline
    TL      & \db{db4} & \db{db1}   &  [011]       & $90.9 \pm 3.5$ & $144.9 \pm 82.9$\\
    TL      & \db{db4} & \db{db1}   &  [001]       & $89.9 \pm 3.6$ & $152.0 \pm 55.7$\\
    \hline
    TL      & \db{db4} & \db{db1}   &  [110]       & $91.8 \pm 3.7$ & $138.7 \pm 89.1$\\
    TL      & \db{db4} & \db{db1}   &  [100]       & $91.0 \pm 4.2$ & $131.0 \pm 84.0$\\
    \hline
    TL      & \db{db4} & \db{db1}   &  [111]       & $91.8 \pm 3.1$ & $158.1 \pm 90.8$\\
  \hline\hline
  \end{tabular}
\end{table*}
Results in \tref{tab:TLdb4} show that all of the \ac{TL} settings improve the baseline result (79\%). The accuracy improvement is at least 10\%.
As mentioned before, the low accuracy rate of the target problem is related to the high variability of the particles and image artifacts.
This causes a severe underperformance on the network.
By learning the source problem dataset \db{db4}, we are thus providing a better starting point for the model to learn the target problem.
In fact, transferring the model and letting all layer weights be fine-tuned ([111]) to the target problem achieved the best result (approx. 92\%).
Another result from our experimental study is concerned with the relevance of the feature representation that the network has captured.
From \tref{tab:TLdb4} we can infer that the first layers are the most relevant. In fact, letting the first two layers of the network be re-learnt and fixing only the last layer (\ac{TL} setting [110]) achieved a higher performance (approx. 92\%) in comparison with only re-learning the last layer ([001] with an accuracy of 90\%).

\begin{table*}[!htp]
  \centering
  \caption{Results of the application of \ac{TL} to the recognition of immunogold particles. The baseline model was trained in a standard \ac{ML} way on the dataset with magnification of \db{15000} (target problem). A model trained for the dataset with magnification of \db{30000} (source problem, immunogold particles were resized) was obtained and reused on the target problem. Overall, all \ac{TL} approaches achieved an improvement of more than 10\%. Each layer-wise \ac{TL} strategy is illustrated in the column \ac{TL} setting (see main text). Results were averaged over the 20 repetitions.}
  \label{tab:TLdb3}
  \begin{tabular}{p{1.5cm}p{1.3cm}p{1.cm}>{\centering\arraybackslash}p{2.5cm}p{2cm}p{3cm}}
    \hline\hline
    Method  & Source (resized)          & Target       & Reusability and Fine-Tuning (TL) Setting & Accuracy ($\pm$~std. dev) & Pre-Training + Fine-Tuning Time (sec.)  \\ \hline
    Baseline&    -     & \db{db1}   &  -           & $79.1 \pm \p5.6$ & $255.1 \pm 83.0$ \\
    \hline
    TL      & \db{db3} & \db{db1}   &  [011]       & $91.4 \pm \p5.9$ & $166.1 \pm 84.8$\\
    TL      & \db{db3} & \db{db1}   &  [001]       & $89.8 \pm \p8.2$ & $119.5 \pm 77.4$\\
    \hline
    TL      & \db{db3} & \db{db1}   &  [110]       & $91.1 \pm \p6.3$ & $152.2 \pm 95.3$\\
    TL      & \db{db3} & \db{db1}   &  [100]       & $87.7 \pm  12.8$ & $121.2 \pm 97.0$\\
    \hline
    TL      & \db{db3} & \db{db1}   &  [111]       & $92.3 \pm \p4.6$ & $220.1 \pm 56.0$\\
  \hline\hline
  \end{tabular}
\end{table*}
A similar experience was conducted when using \db{db3} as source problem. Once again, reusing and fine-tuning the full network (reusability setting [111]) lead to the best results in comparison to the baseline: 13\% improvement of performance accuracy. Such results are depicted in \tref{tab:TLdb3}.
\begin{table*}[!htp]
  \centering
  \caption{Results of the application of \ac{TL} to the recognition of immunogold particles. The baseline model was trained in a standard \ac{ML} way on the dataset with magnification of \db{15000} (target problem). A model trained for the dataset with magnification of \db{20000} was obtained and reused on the target problem. Overall, all \ac{TL} approaches achieved an improvement of more than 10\%. Each layer-wise \ac{TL} strategy is illustrated in the column \ac{TL} setting (see main text). Results were averaged over the 20 repetitions.}
  \label{tab:TLdb2}
  \begin{tabular}{p{1.5cm}p{1.3cm}p{1.cm}>{\centering\arraybackslash}p{2.5cm}p{2cm}p{3cm}}
    \hline\hline
    Method  & Source                & Target       & Reusability and Fine-Tuning (TL) Setting & Accuracy ($\pm$~std. dev) & Pre-Training + Fine-Tuning Time (sec.)  \\ \hline
    Baseline&    -     & \db{db1}   &  -           & $79.1 \pm \p5.6$ & $255.1 \pm \p83.0$ \\
    \hline
    TL      & \db{db2} & \db{db1}   &  [011]       & $92.9 \pm \p2.9$ & $304.3 \pm  129.0$\\
    TL      & \db{db2} & \db{db1}   &  [001]       & $92.4 \pm \p2.9$ & $194.9 \pm  123.6$\\
    \hline
    TL      & \db{db2} & \db{db1}   &  [110]       & $92.9 \pm \p3.3$ & $290.6 \pm  116.5$\\
    TL      & \db{db2} & \db{db1}   &  [100]       & $92.1 \pm \p4.0$ & $251.3 \pm  146.5$\\
    \hline
    TL      & \db{db2} & \db{db1}   &  [111]       & $93.3 \pm \p2.9$ & $326.1 \pm \p78.1$\\
  \hline\hline
  \end{tabular}
\end{table*}
A similar conclusion can be obtained when using the dataset \db{db2} as source problem (see \tref{tab:TLdb2}). Reusing and fine-tuning all layers of the network rendered the best results with 93.3\% of accuracy performance.
Overall, using any \ac{TL} setting for learning the immunogold problem outperformed the baseline (79\% for the dataset \db{db1}).

We have also tested the applicability of the \ac{TL} models over the detections provided by \ac{LoG}. In a first experiment \db{db4} was defined as source problem and \db{db1} as target problem. Once the learning model for the source problem was obtained, we have applied it to the detections provided by the \ac{LoG} filter. Results are depicted in \tref{table:PrecisionRecallResults}.
\begin{table*}[!htp]
\caption{F-measure performance for the best Precision and Recall for LoG and LoG coupled with \ac{SDA} (see \fref{fig:log_sda_results}). Best results are in bold and presented in percentage. Results were averaged over the 20 repetitions.}
\footnotesize
\begin{center}
\begin{tabular}{c|cc|p{2cm}p{2cm}p{2cm}}
  \hline\hline
                   & LoG     & LoG+SDA   & \multicolumn{3}{c}{LoG+SDA by TL [111]} \\
  \cline{4-6}

\hline
                   & -    & -       & Source Problem: \db{db4}  & Source Problem: \db{db3}  & Source Problem: \db{db2}  \\
\hline
\db{db1}           & 84.2 (1.5) & 83.9 (2.3)    & \textbf{85.4 (1.7)}   & \textbf{85.2 (1.9)}  & \textbf{85.4 (1.7)}\\
\hline\hline
\end{tabular}
\end{center}
\normalsize
\label{table:PrecisionRecallResults}
\end{table*}

As mentioned, the reduced training times is one of the many advantages for using \ac{TL} strategies. In particular for \acp{DNN} the gains can be significantly high due to the time that it takes in such deep (big) networks.
Training an \ac{SDA} for the baseline (target) problem (i.e., dataset \db{db1}) took on average 255 seconds. When using \ac{TL} with the dataset \db{db4} as source problem, the training time is reduced to 158 seconds (reusing and fine-tuning all layers of the network --- see \tref{tab:TLdb4}). In other words, we obtain a reduction of 38\% on the training time. This indeed shows the advantage on using \ac{TL} as a way to speed-up the training time.
When using the dataset \db{db3} as source problem and, reusing and fine-tuning all layers of the network (see \tref{tab:TLdb3}) for the dataset \db{db1} takes in average 220 seconds.
In short, if one wishes to solve the \db{db1} one ought use the \db{db4} as source problem. Solving \db{db1} by \ac{TL} will take 239.8 seconds corresponding to the total time of training the baseline model of \db{db4} plus the \ac{TL} training time. It still takes less time than training the \db{db1} baseline model and with an increased accuracy performance improvement.

\section{Conclusions}
\label{sec:conclusions}

In this work we proposed a framework for the automatic detection of immunogold particles in different magnifications. We found that solely the \ac{LoG} filter attained results of over 84\% of accuracy with the F-measure.
Regarding \ac{TL} for immunogold recognition, the conclusions of our study are two-fold: a) there was a significant accuracy improvement when applying \ac{TL} (around 10\% of improvement); and, b) a reduction in training times. Overall, to solve the problem of immunogold recognition on low magnifications (i.e., magnification of \db{15000}, \db{db1}) it is better to use as source problem the dataset \db{db4}, magnification \db{50000}.

These results show that these approaches are also resilient to the presence of noise, artifacts and cluttered background and are easy to set-up based on the few parameters of the framework (only dependent on the threshold parameter of the \ac{LoG} filter response).
These methods can help a bio-researcher in the tedious and time-consuming task of identifying immunogold particles in large EM images.

We also make available the developed code and the dataset used such that our work can be reproduced and improved by the community.

\section{Acknowledgements}
This work was financed by FEDER funds through the Programa Operacional Factores de Competitividade – COMPETE and by Portuguese funds through FCT – in the framework of the project PTDC/EIA-EIA/119004\\/2010 and UID/BIM/04293/2013.
Tiago Esteves also acknowledges FCT through fellowship grant number SFRH/BD/80508/2011.
Sara Rocha was supported by Grant BIIC M3.1.6/F/038/2009 from Direcção Regional de Ciência e Tecnologia and by Grant SFRH/BD/8122/2002 from FCT.
We thank Roberto Salema for his insightful comments, to Rui Fernandes from HEMS department, and to João Relvas for aiding us with the technical knowledge for conducting this work.
%

\end{document}